\documentclass{article}
\PassOptionsToPackage{numbers,sort&compress}{natbib}
\usepackage[final,main]{neurips_2026}
\usepackage[utf8]{inputenc} 
\usepackage[T1]{fontenc}    
\usepackage{hyperref}       
\usepackage{url}            
\usepackage{booktabs}       
\usepackage{amsfonts}       
\usepackage{nicefrac}       
\usepackage{microtype}      
\usepackage{xcolor}         
\usepackage{graphicx} 
\usepackage{amsmath}
\usepackage{multirow}
\usepackage{pifont}
\usepackage{wrapfig}

\title{Complexity-guided Regularization for Generalizable Human Gaussian Splatting}

\author{%
  Gun Ryu$^{1}$ \quad Gi-Mun Um$^{2}$ \quad Won-Sik Cheong$^{2}$ \quad Wonjun Kim$^{1}$\thanks{Corresponding author.} \\[2pt]
  $^{1}$Konkuk University \\
  $^{2}$Electronics and Telecommunications Research Institute \\
  \texttt{\{fbrjs15, wonjkim\}@konkuk.ac.kr} \quad \texttt{\{gmum, wscheong\}@etri.re.kr}
}

\begin{document}

\maketitle
\setcounter{footnote}{0}

\begin{abstract}
Recent studies on generalizable human Gaussian splatting have achieved the notable progress in synthesizing novel views of unseen human subjects from multi-view images. 
Despite this progress, existing methods still struggle to represent fine-grained details of human subjects, which inherently involve high geometric and textural complexity. 
This is because previous methods do not impose a constraint on how densely Gaussians are placed, which often leads to blurry rendering results in complex regions. 
To address this problem, we propose a complexity-guided regularization scheme. The key idea is to leverage a complexity prior computed via the local entropy in geometric and textural characteristics of a clothed human. 
This effectively prevents Gaussians from being coarsely placed in complex regions, which facilitates the faithful reconstruction of intricate structures. 
Furthermore, we sample local patches predominantly from complex regions according to this complexity prior, which allows the model to be sufficiently supervised by local details. Experimental results on benchmark datasets demonstrate that our method delivers state-of-the-art performance while maintaining the real-time inference speed in generalizable human Gaussian splatting. The code and model are publicly available at link\footnote{\url{https://github.com/DCVL-3D/CGR-GS_release}}.

\end{abstract}

\section{Introduction}
Realistic 3D human reconstruction has gained the significant interest due to its wide range of applications in virtual reality, telepresence, gaming, content creation, etc. 
In recent literature, 3D Gaussian Splatting (3DGS)~\cite{3dgs}, which explicitly represents a scene as a set of 3D Gaussian primitives while enabling real-time rendering, has been widely adopted for this task.
Most recently, various studies~\cite{hugs,gart,3dgsavatar,gomavatar,animatable3dgs,splattingavatar,gauhuman} have begun to integrate human body templates, e.g., SMPL~\cite{smpl}, to improve the rendering performance.
However, since previous methods rely on the per-subject optimization scheme, they cannot be directly applied to unseen individuals without repeating the entire training procedure, which severely restricts their practicality in real-world scenarios.

To overcome the aforementioned limitation, recent studies have explored generalizable approaches, which regress Gaussian attributes for unseen human subjects in a single forward pass. Due to the absence of the iterative refinement process, the majority of generalizable methods rely on an explicit structural prior to mitigate the difficulty of directly predicting Gaussian attributes.
For example, some methods~\cite{gps-gaussian, eva-gaussian, mvsc-ghgs} have exploited geometric constraints between view pairs as a structural prior to determine the spatial layout of Gaussians.
A skinned human model has also been employed~\cite{ghg, lifegom, rogsplat} to enable the feed-forward model to place Gaussians directly on the body surface.
Even though considerable efforts have been devoted to determining where to place Gaussians via feed-forward approaches, how densely they should be placed across different regions of the human subject has not been sufficiently explored.

\begin{figure*}[t]
    \centering
    \includegraphics[width=\textwidth]{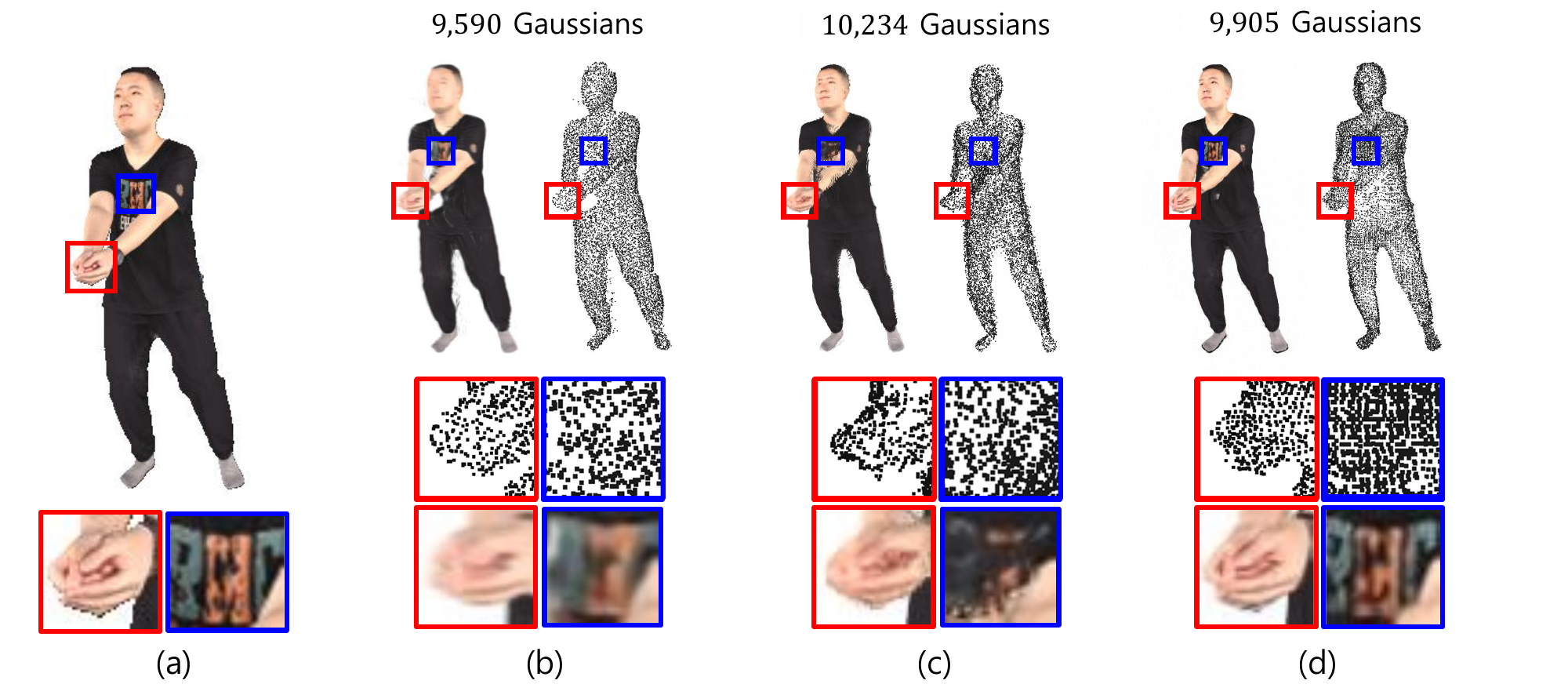}
    \caption{\textbf{Distributions of Gaussian centers and the corresponding rendering results.} (a) Novel view image. (b) GPS-Gaussian~\cite{gps-gaussian}. (c) RoGSplat~\cite{rogsplat}. (d) Proposed method.}
    \label{fig:fig1}
\end{figure*}

In this paper, we introduce a novel approach for effectively handling generalizable human Gaussian splatting.
Due to the inherent smoothness of Gaussian primitives, Gaussians need to be densely placed to faithfully reconstruct high-frequency details.
However, previous feed-forward methods localize Gaussians based on the pixel grid~\cite{gps-gaussian, eva-gaussian, mvsc-ghgs} or vertices of the skinned mesh~\cite{rogsplat, ghg, lifegom}, whose density is highly related to the image resolution or the skinned mesh template rather than the visual properties of the target subject itself.
While this may not be significant in simple regions (e.g., plain areas), it often leads to an insufficient number of Gaussians in complex regions (e.g., fingers or textured clothes), which produces blurry rendering results as shown in Figs.~\ref{fig:fig1}(b) and (c). 
To deal with this problem, we propose a complexity-guided regularization scheme.
The key idea is to allocate Gaussians in proportion to the local complexity of the clothed human mesh.
Specifically, we define the complexity prior as the entropy of the local geometric and textural information around each vertex of the human mesh.
Guided by this prior, we regularize the local density of Gaussians to be positively correlated with the complexity.
This helps the model faithfully reconstruct high-frequency details with sufficient Gaussians, as shown in Fig.~\ref{fig:fig1}(d).
Moreover, we find that supervising the model with the entire image is suboptimal for reconstructing fine-grained details. 
This is because supervisory signals from complex regions, which often occupy only a small portion of the human-centric image, e.g., fingers, facial features, etc., are easily dominated by those from surrounding simple regions.
Accordingly, during the training phase, we use the complexity prior as sampling weights to select local patches, where regions of higher complexity are more frequently sampled.
This training strategy enables the model to receive sufficient supervision on fine details.
The main contribution of the proposed method can be summarized as follows:
\begin{itemize}
\item We define a complexity prior as a per-vertex scalar value, which is computed via the local entropy in geometric and textural characteristics of the clothed human mesh.
Leveraging this prior, we propose a novel regularization scheme to guide the model to adjust the local Gaussian density according to the complexity.
This encourages the model to allocate Gaussians densely in complex regions, which enables faithful reconstruction of fine-grained details.
\item To effectively provide sufficient supervisory signals on intricate details, we introduce a patch-based training strategy, which samples local patches with probability proportional to the complexity prior.
\end{itemize}



\section{Related work}

\noindent\textbf{Per-subject optimization-based 3DGS for human rendering.}
Motivated by the great success of 3D Gaussian splatting~\cite{3dgs} in novel view synthesis, several studies have begun to apply 3DGS to human rendering based on the point-based representation. 
For example, Gaussian points are initialized on faces of the human mesh and optimized to represent the detailed appearance while preserving the geometric structure of the target subject~\cite{gomavatar, splattingavatar}.
On the other hand, other methods~\cite{gaussianavatar, 3dgsavatar, gart, hugs, animatable3dgs, gauhuman} optimize Gaussian attributes to reconstruct the human in a canonical pose, which are rendered in arbitrary poses via forward skinning.
Although these approaches have demonstrated the impressive performance, they typically require the per-subject optimization, which restricts their generalizability to unseen human subjects.

\noindent\textbf{Generalizable 3DGS for human rendering.}
To overcome this limitation, recent studies have explored generalizable human Gaussian splatting, which regress Gaussian attributes for unseen subjects in a single forward pass.
As a pioneer, Zheng \textit{et al.}~\cite{gps-gaussian} estimated per-pixel Gaussian attributes based on the depth map, which is predicted through stereo matching between adjacent views.
In a similar vein, Hu \textit{et al.}~\cite{eva-gaussian} improved the regression process by employing the 1D-window attention along epipolar lines between view pairs.
More recently, Kim \textit{et al.}~\cite{mvsc-ghgs} leveraged a foundation model~\cite{vggt} to inject the implicit 3D geometric knowledge, while recalibrating multi-view features via semantic consistency-based cross-view attention, which helps the model robustly estimate Gaussian attributes in visually complex regions.
Instead of estimating per-pixel Gaussian attributes, skinned human models have been adopted as a structural prior to directly place Gaussians on the body surface.
In this direction, Wen \textit{et al.}~\cite{lifegom} attached Gaussians to the skinned mesh surface and iteratively refined Gaussian attributes through a feedback mechanism.
Similarly, Xiao \textit{et al.}~\cite{rogsplat} densified SMPL vertices to align with the actual human geometry and predicted Gaussian attributes on such densified points.
Kwon \textit{et al.}~\cite{ghg} regressed Gaussian attributes in the 2D UV space of the SMPL template and introduced a multi-scaffold representation to effectively mitigate the geometric discrepancy between the skinned body and clothed human.
While previous efforts have focused on accurately estimating the position of Gaussians, we shift our focus to how densely Gaussians should be placed and propose an effective regularization scheme, which significantly improves the rendering performance of generalizable human Gaussian splatting.



\section{Method}

\begin{figure*}[t]
    \centering
    \includegraphics[width=0.95\textwidth]{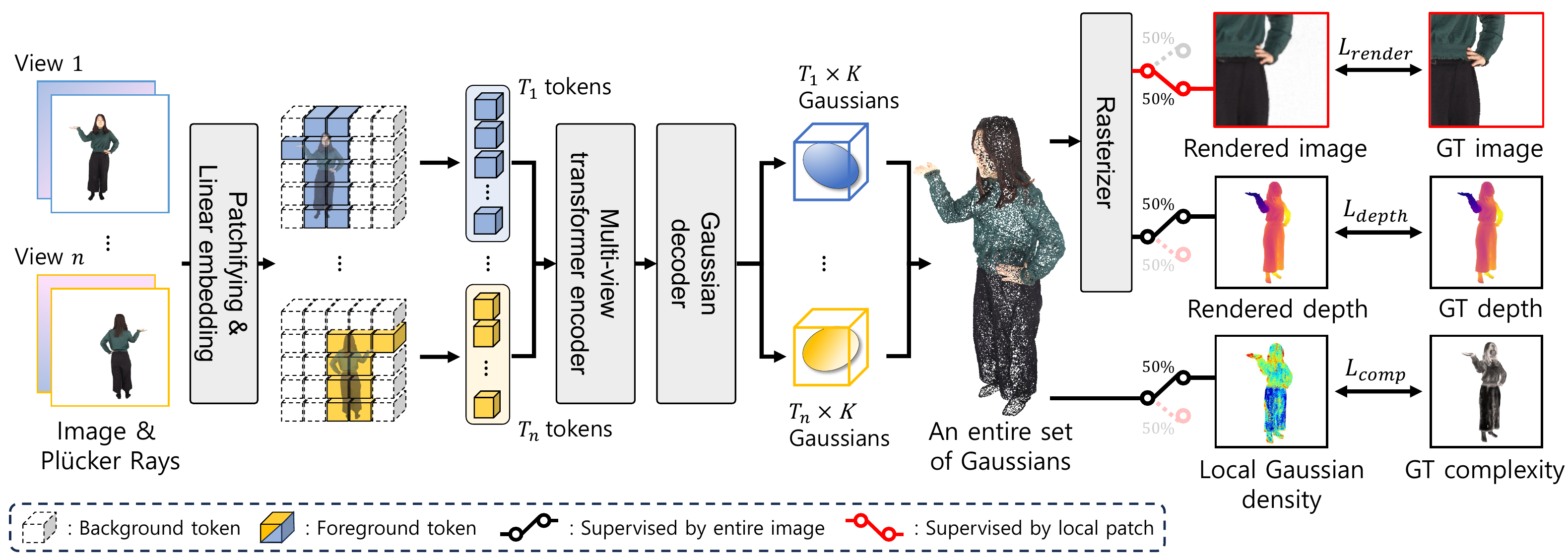}
    \caption{\textbf{Overall architecture of the proposed method.} The complexity prior of the clothed human guides the model to allocate Gaussians densely in complex regions. Note that Gaussians are randomly supervised by either the entire image or complexity-aware local patches with the equal probability at each iteration.}
    \label{fig:overall}
\end{figure*}

\subsection{Architecture details}
The overall architecture of the proposed method is shown in Fig.~\ref{fig:overall}.
Given $N$ multi-view RGB images $\{\mathbf{I}_n\}_{n=1}^{N} \in \mathbb{R}^{3 \times H \times W}$ of a clothed human subject with the corresponding Pl\"{u}cker ray~\cite{plucker} maps $\{\mathbf{P}_n\}_{n=1}^{N} \in \mathbb{R}^{6 \times H \times W}$ and human masks $\{\mathbf{M}_n\}_{n=1}^{N} \in \{0,1\}^{H \times W}$, our goal is to regress a set of Gaussian attributes in a single feed-forward pass.
Following previous feed-forward approaches~\cite{gs-lrm, grm}, each input image is concatenated with the corresponding Pl\"{u}cker ray map along the channel dimension, which yields a per-view feature map $\tilde{\mathbf{I}}_n = [\mathbf{I}_n ; \mathbf{P}_n] \in \mathbb{R}^{9 \times H \times W}$.
The resulting feature map $\tilde{\mathbf{I}}_n$ is partitioned into non-overlapping $p\times p$ patches, which are subsequently flattened and projected into tokens.
Since the target subject only occupies a small area of the human-centric image, we discard tokens whose patch lies entirely outside the human mask $\mathbf{M}_n$, which yields $T_n$ foreground tokens for the $n$-th view.
This significantly reduces the number of tokens fed into transformer layers, which leads to the real-time inference.
Following Long-LRM++~\cite{longlrm++}, a Gaussian decoder regresses attributes of $K$ Gaussians from each token.
Specifically, the decoder predicts depth $d\in\mathbb{R}$, 3D spatial offset $\Delta\mathbf{x}\in\mathbb{R}^{3}$, scale $\mathbf{s}\in\mathbb{R}^{3}$, rotation quaternion $\mathbf{q}\in\mathbb{R}^{4}$, opacity $o\in\mathbb{R}$, and color $\boldsymbol{\rho}\in\mathbb{R}^{3}$.
The 3D position of each Gaussian is computed by unprojecting the predicted depth $d$ along the ray as $\mathbf{r}_o+d\,\mathbf{r}_d$, which is further shifted by the spatial offset $\Delta\mathbf{x}$. Note that $\mathbf{r}_o\in\mathbb{R}^{3}$ and $\mathbf{r}_d\in\mathbb{R}^{3}$ denote directional vectors of the origin and the center ray of the patch, respectively.
Consequently, the model produces $K \times T$ Gaussians, where $T = \sum_{n=1}^{N} T_n$ is the total number of foreground tokens across all views, which are rasterized to synthesize the target view image.

\subsection{Complexity prior of the human subject}
A clothed human is inherently a complex subject whose surface contains structural and visual information at a wide range of spatial scales.
Moreover, such complexity varies significantly across different regions of the surface.
Properly accounting for this variation is essential to faithfully reconstruct the human subject.
To this end, we quantitatively define the local complexity of a clothed human, which can serve as an explicit guidance signal for the reconstruction process.
Specifically, we represent the clothed human as a textured mesh $\mathcal{M}=(\mathbf{V},\mathbf{F},T)$, whose components are defined as follows:
$\mathbf{V}=\{\mathbf{v}_i\}_{i=1}^{|\mathbf{V}|}$ is the set of vertices, where $\mathbf{v}_i\in\mathbb{R}^{3}$ represents a point in the 3D space.
$\mathbf{F}=\{\mathbf{f}_k\}_{k=1}^{|\mathbf{F}|}$ is the set of triangular faces, where $\mathbf{f}_k=(\mathbf{v}_{k_1}, \mathbf{v}_{k_2}, \mathbf{v}_{k_3})$ denotes a triplet of vertices which constitute a face.
$T \in \mathbb{R}^{H \times W \times 3}$ is a UV map for the human mesh, which stores RGB color values.
Each pixel in $T$ corresponds to a point on the mesh surface via UV coordinates.
On this mesh, we define the complexity prior as a scalar value $c_i\in[0,1]$ for vertex $\mathbf{v}_i$.

\noindent\textbf{Geometric complexity.}
We denote the set of faces including the vertex $\mathbf{v}_i$ as $\mathcal{F}_i$.
For each vertex $\mathbf{v}_i$, we compute an area-weighted unit normal $\mathbf{n}_i$, which represents the dominant normal direction of the local surface as follows:
\begin{equation}
    \mathbf{n}_i=\frac{\mathbf{m}_i}{\|\mathbf{m}_i\|},\quad\text{where}\quad\mathbf{m}_i=\sum_{\mathbf{f}_k\in\mathcal{F}_i}A_{\mathbf{f}_k}\,\mathbf{n}_{\mathbf{f}_k},
    \label{eq:ref_normal}
\end{equation}
where $\mathbf{n}_{\mathbf{f}_k}\in\mathbb{R}^{3}$ and $A_{\mathbf{f}_k}\in\mathbb{R}_{>0}$ denote the unit normal and the area of face $\mathbf{f}_k$, respectively.
We further find a local neighborhood $\mathcal{N}_i=\{\mathbf{v}_j : j\ne i,\; \|\mathbf{v}_j-\mathbf{v}_i\|\le r\}$ of $\mathbf{v}_i$ within a radius $r$ to measure how the unit normal varies around it.
For neighbor $\mathbf{v}_j\in\mathcal{N}_i$, the angular deviation of its area-weighted unit normal $\mathbf{n}_j$ from $\mathbf{n}_i$ is defined as $s^{\text{geo}}_j=(1-\mathbf{n}_i\cdot\mathbf{n}_j)/2\in[0,1]$.
Subsequently, we construct a histogram $\mathbf{p}^{\text{geo}}_i=(p^{\text{geo}}_1,\ldots,p^{\text{geo}}_B)$ of the deviations $\{s^{\text{geo}}_j\}_{j\in\mathcal{N}_i}$ over $B$ equal-width bins on $[0,1]$, where $p^{\text{geo}}_b$ denotes the proportion of deviations falling into bin $b$.
The geometric complexity at $\mathbf{v}_i$ is finally computed by the normalized Shannon entropy~\cite{shannon} of this histogram, as follows:
\begin{equation}
    H^{\text{geo}}_i=-\frac{1}{\log_2 B}\sum_{b=1}^{B} p^{\text{geo}}_b\log_2 p^{\text{geo}}_b.
    \label{eq:geo_entropy}
\end{equation}

\begin{figure*}[t]
    \centering
    \includegraphics[width=0.9\textwidth]{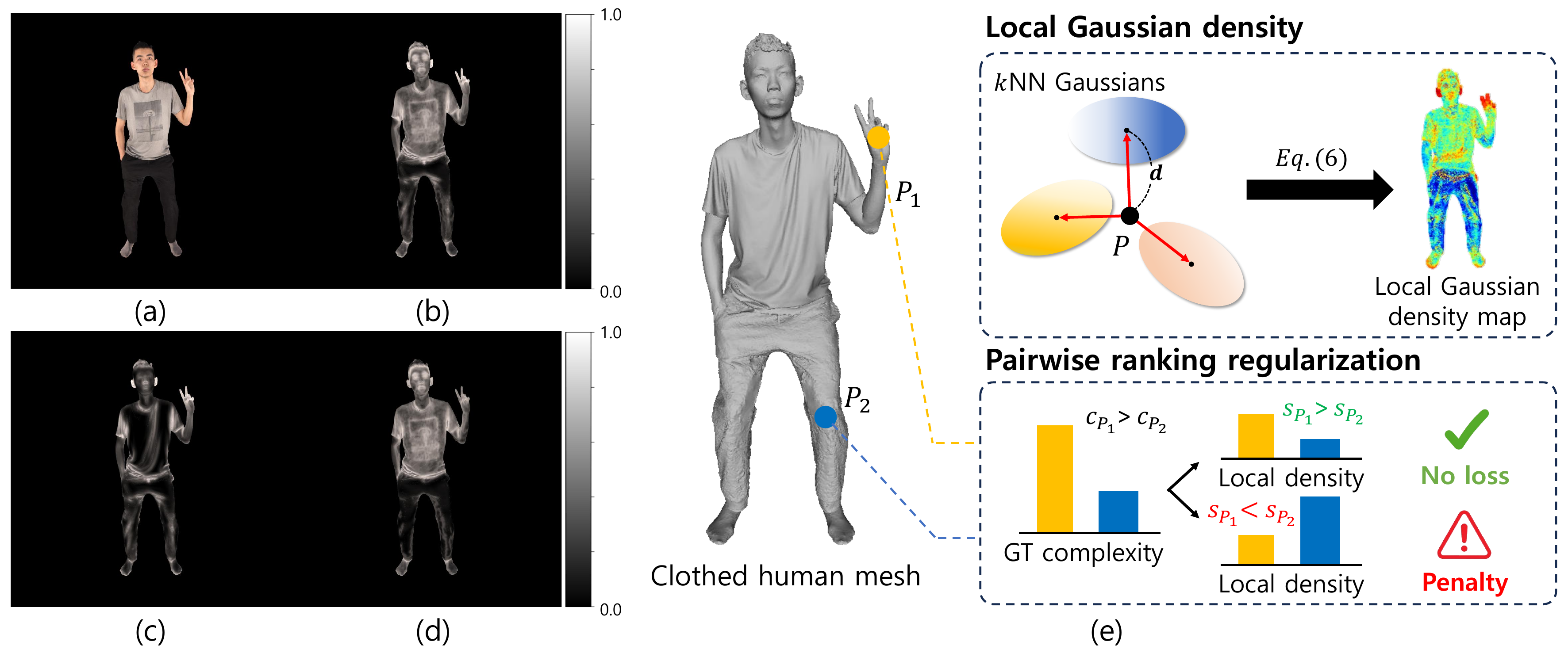}
    \caption{\textbf{Complexity prior and complexity-guided regularization scheme.} (a) RGB image. (b) Per-vertex complexity prior $c$. (c) Geometric complexity $\bar{H}^{\text{geo}}$. (d) Textural complexity $\bar{H}^{\text{tex}}$. (e) Overview of the complexity-guided regularization process, where the local Gaussian density is computed from the average distance of $k$-nearest neighbor Gaussians and regularized to be positively correlated with the complexity prior via a pairwise ranking regularization scheme.}
    \label{fig:complexity}
\end{figure*}

\noindent\textbf{Textural complexity.}
Similarly, we compute the textural complexity around each vertex.
The RGB color $\mathbf{t}_i$ of the vertex $\mathbf{v}_i$ is retrieved from the location $(u_i, v_i)$ of the corresponding UV map $T$.
For $\mathbf{v}_j\in\mathcal{N}_i$, the chromatic deviation of its color $\mathbf{t}_j$ from the color $\mathbf{t}_i$ is defined as $s^{\text{tex}}_j = \|\mathbf{t}_j - \mathbf{t}_i\|/\sqrt{3} \in [0,1]$.
Subsequently, we construct a histogram $\mathbf{p}^{\text{tex}}_i = (p^{\text{tex}}_1, \ldots, p^{\text{tex}}_B)$ of the deviations $\{s^{\text{tex}}_j\}_{j\in\mathcal{N}_i}$ over $B$ equal-width bins on $[0,1]$, where $p^{\text{tex}}_b$ denotes the proportion of deviations falling into bin $b$.
The textural complexity at $\mathbf{v}_i$ is also computed by the normalized Shannon entropy~\cite{shannon} of this histogram, as follows:
\begin{equation}
    H^{\text{tex}}_i=-\frac{1}{\log_2 B}\sum_{b=1}^{B} p^{\text{tex}}_b\log_2 p^{\text{tex}}_b.
    \label{eq:tex_entropy}
\end{equation}

\noindent\textbf{Multi-scale aggregation.}
To account for complexity at multiple spatial scales, we repeat above computations at $S$ different scales with radii $\{r_s\}_{s=1}^{S}$, where $r_s$ is set to a different fraction of the bounding box diagonal of $\mathcal{M}$.
Entropies $H^{\text{geo}}_{i,s}$ and $H^{\text{tex}}_{i,s}$ at scale $s$ are  averaged to obtain the multi-scale complexities, as follows: $\bar{H}^{\text{geo}}_i = \frac{1}{S}\sum_{s=1}^{S} H^{\text{geo}}_{i,s}$ and $\bar{H}^{\text{tex}}_i = \frac{1}{S}\sum_{s=1}^{S} H^{\text{tex}}_{i,s}$, respectively.
Consequently, the per-vertex complexity prior $c_i$ is finally defined as the maximum between the geometric and textural complexities as follows:
\begin{equation}
    c_i = \max\!\big(\bar{H}^{\text{geo}}_i,\;\bar{H}^{\text{tex}}_i\big).
    \label{eq:complexity_prior}
\end{equation}
This maximum operation ensures a high complexity value whenever at least one of these cues shows a high degree of local variations.
As shown in Figs.~\ref{fig:complexity}(b), (c), and (d), the complexity prior effectively captures the local variation in both geometric and textural characteristics of the clothed human, where higher values are attributed to structurally or visually intricate regions (e.g., fingers or textured clothes).
The resulting complexity prior $\{c_i\}$ serves as the guidance signal for regularization and patch-based supervision, which are explained in the following subsections.

\subsection{Complexity-guided regularization}
Due to the inherent smoothness of Gaussian primitives, the local density of Gaussians should be properly taken into account for faithfully reconstructing fine-grained details.
However, previous methods allocate Gaussians without regard to the local complexity of the human subject, which leaves fine-grained regions under-represented.
To address this problem, we introduce a regularization scheme, which couples the local Gaussian density with the complexity prior $\{c_i\}$ and explicitly guides the model to allocate Gaussians densely in complex regions.
An overview of this scheme is shown in Fig.~\ref{fig:complexity}(e).
Since the complexity prior is defined at mesh vertices, we first sample a set of points on the surface of the human mesh.
Specifically, we randomly sample $M$ points on the mesh surface with the probability proportional to the area of the corresponding face $\mathbf{f}_m$.
The complexity value $c_m\in[0,1]$ at sampled point $\mathbf{p}_m\in\mathbb{R}^{3}$ of face $\mathbf{f}_m$ is obtained by interpolating the complexity values at three vertices $\{\mathbf{v}_j\}_{j\in\mathbf{f}_m}$ as follows:
\begin{equation}
    c_m = \sum_{j\in\mathbf{f}_m} \alpha_j\, c_j,
\end{equation}
where $\{\alpha_j\}_{j\in\mathbf{f}_m}$ are the barycentric weights of $\mathbf{p}_m$, which satisfies $\mathbf{p}_m = \sum_{j\in\mathbf{f}_m} \alpha_j\, \mathbf{v}_j$ and $\sum_{j\in\mathbf{f}_m} \alpha_j = 1$.
Following CDC-GS~\cite{cdc-gs}, we measure the local Gaussian density at sampled point $\mathbf{p}_m$ based on the distances to the $k$-nearest Gaussian centers.
Note that Gaussians whose opacity is below 0.1 are excluded from the nearest-neighbor search.
Specifically, the local density score $s_m$ is defined as follows:
\begin{equation}
    s_m = -\log\!\left(\frac{1}{k}\sum_{j=1}^{k} d_{m,j} + \epsilon\right),
    \label{eq:knn_density}
\end{equation}
where $d_{m,j}$ denotes the Euclidean distance from $\mathbf{p}_m$ to its $j$-th nearest Gaussian center and $\epsilon$ is a small positive constant for numerical stability.
To encourage $s_m$ to be positively correlated with $c_m$, we adopt a pairwise monotonic ranking regularization scheme.
Given a pair of points on the mesh surface $(\mathbf{p}_a,\mathbf{p}_b)$ with $c_a > c_b$, the regularization penalizes the case in which the point with higher complexity has a lower density score, i.e., $s_a<s_b$.
The complexity-guided regularization is formulated as follows:
\begin{equation}
    \mathcal{L}_{\text{comp}} = \frac{1}{|\mathcal{P}|}\sum_{(a,b)\in\mathcal{P}} (c_a - c_b)\cdot\text{softplus}\!\left(-(s_a - s_b)\right),
    \label{eq:rank_loss}
\end{equation}

where $\mathcal{P}$ denotes the set of sampled pairs $(a,b)$
and $\text{softplus}(x) = \ln(1 + e^x)$. Each pair is weighted by the complexity gap $c_a - c_b$ to impose a strong penalty on pairs with a significant complexity difference.
This regularization scheme encourages the model to place Gaussians densely in complex regions, which enables the faithful reconstruction of fine-grained details.

\subsection{Complexity-aware patch-based supervision}
Complex regions often occupy only a small portion of the human-centric image, which makes their supervisory signals easily dominated by those from surrounding simple regions during training.
To address this issue, we introduce a patch-based supervision strategy which leverages the complexity prior to concentrate the supervisory signal on intricate regions.
For target view $q$, we first render a 2D complexity map $\mathbf{C}^{\text{2D}}_q\in[0,1]^{H\times W}$ by rasterizing the per-vertex complexity values $\{c_i\}$ onto the image plane.
From this map, we sample $N_p$ patch centers per view with the per-pixel sampling probability given by $\mathbf{C}^{\text{2D}}_q$.
Based on sampled centers, $p_s \times p_s$ local patches are constructed for the supervision.
During the training phase, the model is supervised on either the entire image or these local patches with the equal probability at each iteration, as shown in Fig.~\ref{fig:overall}.
We further adopt a progressive schedule for the size of the patch $p_s$, which gradually decreases from $p_{\text{max}}$ to $p_{\text{min}}$ during training.
This provides a coarse-to-fine supervisory signal, where larger patches in the early stages capture broader structural context and smaller patches in later stages focus on fine-grained details.

\subsection{Loss function}

The proposed method is trained based on three terms, i.e., the rendering loss $\mathcal{L}_{\text{render}}$, the depth regularization $\mathcal{L}_{\text{depth}}$, and the proposed complexity-guided regularization $\mathcal{L}_{\text{comp}}$.
Following~\cite{grm}, the rendering loss $\mathcal{L}_{\text{render}}$ consists of a photometric loss  $\mathcal{L}_{1}$, a perceptual loss $\mathcal{L}_{\text{lpips}}$, and an alpha regularization $\mathcal{L}_{\alpha}$, which aligns the rendered alpha map with the ground-truth human mask.
The depth regularization $\mathcal{L}_{\text{depth}}$ enforces the consistency between rendered and ground-truth depth maps~\cite{gps-gaussian}.
The overall training objective is formulated as follows:
\begin{equation}
\mathcal{L}_{\text{total}} = \lambda_{1} \mathcal{L}_{1} + \lambda_{\text{lpips}} \mathcal{L}_{\text{lpips}} + \lambda_{\alpha} \mathcal{L}_{\alpha} + \lambda_{\text{depth}} \mathcal{L}_{\text{depth}} + \lambda_{\text{comp}} \mathcal{L}_{\text{comp}},
\end{equation}
where $\lambda_{1}$, $\lambda_{\text{lpips}}$, $\lambda_{\alpha}$, $\lambda_{\text{depth}}$, and $\lambda_{\text{comp}}$ are balancing factors for each loss term, which are set to 2.0, 0.5, 0.25, 0.25, and 0.1, respectively.
Note that the same training objective is applied when patch-based supervision, which is explained in the previous subsection, is activated.
\begin{figure*}[t]
    \centering
    \includegraphics[width=0.85\textwidth]{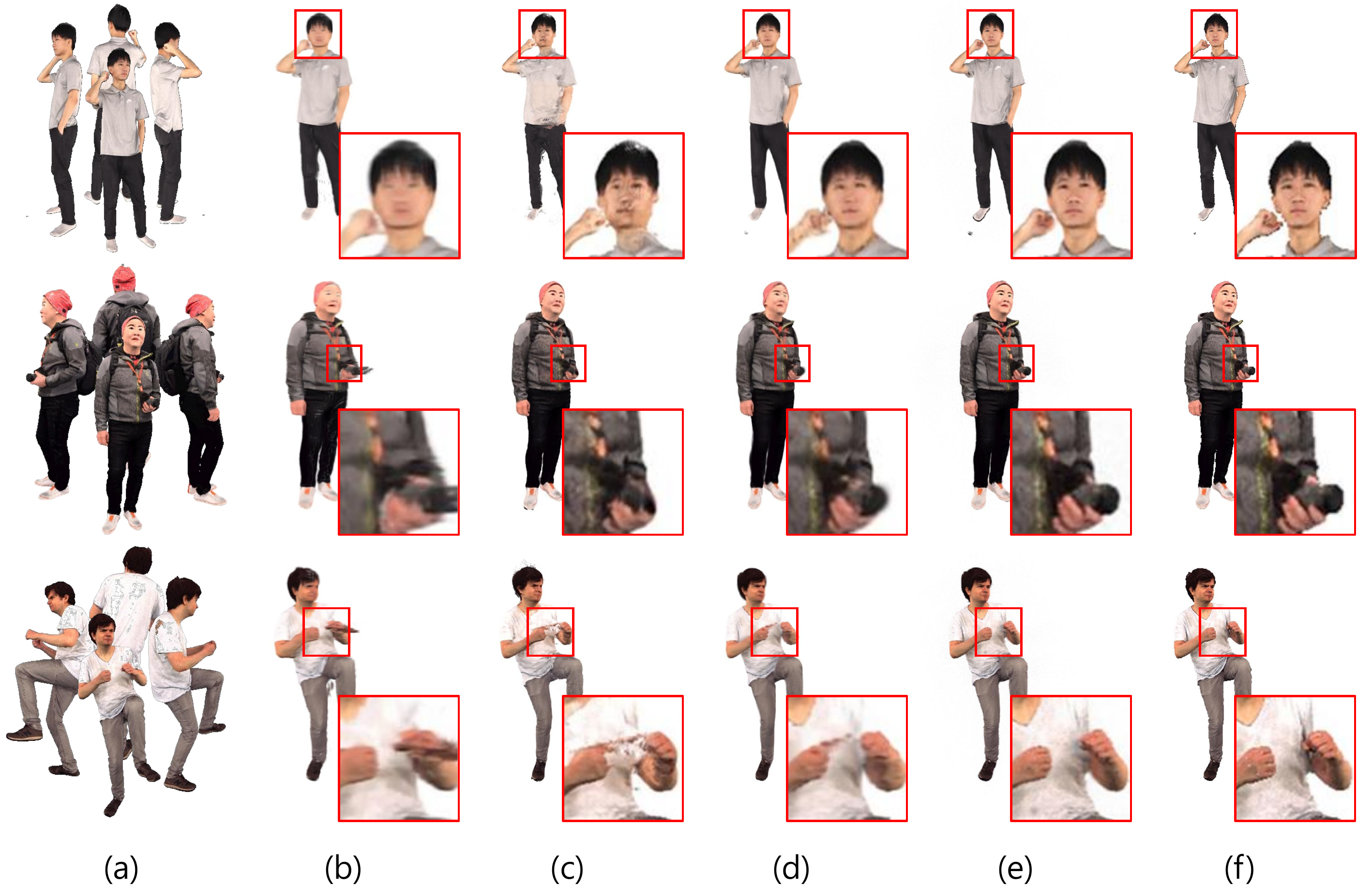}
    \caption{\textbf{Qualitative comparison of in-domain novel view synthesis on THuman2.0~\cite{thuman}, 2K2K~\cite{2k2k}, and X-Humans~\cite{xhuman} datasets (from top to bottom).} (a) Input images. (b) Results by GPS-Gaussian~\cite{gps-gaussian}. (c) Results by RoGSplat~\cite{rogsplat}. (d) Results by GHGS-MVSC~\cite{mvsc-ghgs}. (e) Results by the proposed method. (f) Ground truth.}
    \label{fig:fig4}
\end{figure*}

\section{Experiments}
\subsection{Implementation details}
The proposed method is implemented on the PyTorch~\cite{pytorch} framework. All the experiments were performed on a single PC with an Intel Xeon w5-3433@2.0 GHz CPU and an NVIDIA GeForce RTX 4090 GPU. By following~\cite{gs-lrm}, all the parameters of the proposed network are updated by the AdamW optimizer, where the momentum factors are set to 0.9 and 0.95, respectively. 
The learning rate is set to $2\times10^{-4}$. 
Each input feature map is tokenized with a non-overlapping patch size of 8, and the Gaussian decoder predicts 64 Gaussians per token. 
For computing the complexity prior, the number of histogram bins $B$ is set to 20. 
The complexity prior is computed at three different radii $\{r_s\}_{s=1}^{3}$. Each radius is defined as $r_s = \alpha_s \cdot d_{\text{bbox}}$, where $d_{\text{bbox}}$ denotes the diagonal length of the bounding box for the clothed human mesh and $\alpha_1$, $\alpha_2$, and $\alpha_3$ are set to 0.02, 0.01, and 0.005, respectively.
For the complexity-guided regularization, the number of nearest neighbors $k$ is set to 7. 
At each iteration, 2048 points are sampled on the mesh surface and 4096 pairs are sampled for pairwise ranking regularization. 
For patch-based supervision, the number of patch centers per target image $N_p$ is set to 2.
The patch size is linearly decreased from 512 to 64 over 600{,}000 iterations.

\begin{table}[!t]
\centering
\caption{\textbf{Quantitative comparison of in-domain novel view synthesis on THuman2.0~\cite{thuman}, 2K2K~\cite{2k2k}, and X-Humans~\cite{xhuman} datasets.} The best result is shown in bold.}
\label{tab:in_domain_others}
\footnotesize
\setlength{\tabcolsep}{4pt}
\begin{tabular}{lcccccccccc}
\toprule
\multirow{2}{*}{Methods} & \multicolumn{3}{c}{THuman2.0} & \multicolumn{3}{c}{2K2K} & \multicolumn{3}{c}{X-Humans} & \multirow{2}{*}{FPS$\uparrow$}\\
\cmidrule(lr){2-4}\cmidrule(lr){5-7}\cmidrule(lr){8-10}
& PSNR$\uparrow$ & SSIM$\uparrow$ & LPIPS$\downarrow$ & PSNR$\uparrow$ & SSIM$\uparrow$ & LPIPS$\downarrow$ & PSNR$\uparrow$ & SSIM$\uparrow$ & LPIPS$\downarrow$ & \\
\midrule
GPS-Gaussian~\cite{gps-gaussian} & 26.54 & 0.9539 & 0.0610 & 26.46 & 0.9483 & 0.0611 & 26.29 & 0.9458 & 0.0687 & \textbf{56.31} \\
RoGSplat~\cite{rogsplat} & 28.83 & 0.9615 & 0.0429 & 30.67 & 0.9697 & 0.0310 & 30.31 & 0.9656 & 0.0371 & 5.42 \\
GHGS-MVSC~\cite{mvsc-ghgs} & 30.93 & 0.9710 & 0.0334 & 30.45 & 0.9678 & 0.0326 & 30.60 & 0.9668 & 0.0391 & 2.79 \\
\textbf{Ours} & \textbf{32.42} & \textbf{0.9766} & \textbf{0.0201} & \textbf{31.68} & \textbf{0.9717} & \textbf{0.0229} & \textbf{31.96} & \textbf{0.9703} & \textbf{0.0230} & 33.96 \\
\bottomrule
\end{tabular}
\end{table}

\subsection{Datasets and evaluation metrics}
\noindent\textbf{Datasets.} For the performance evaluation of the proposed method, three human-centric multi-view datasets, i.e., THuman2.0~\cite{thuman}, 2K2K~\cite{2k2k}, and X-Humans~\cite{xhuman}, are adopted.
All three datasets provide high-quality textured 3D human scans, which are rendered into multi-view images at the resolution of 512$\times$512 pixels under known camera parameters.
For THuman2.0, we follow train and test splits adopted in prior works~\cite{rogsplat, ghg, mvsc-ghgs}.
For 2K2K, we follow the split adopted in HART~\cite{hart}.
For X-Humans, the official split is used.
The proposed method is trained on each dataset independently for in-domain evaluations.
We further evaluate the cross-domain generalization ability of our method by training on THuman2.0 and directly testing on 2K2K and X-Humans datasets without any fine-tuning. We additionally provide experimental results on real-world datasets (i.e., DNA-Rendering~\cite{dna} and ZJU-MoCap~\cite{zju}) in the appendix.

\noindent\textbf{Evaluation metrics.}
Following previous works~\cite{rogsplat, ghg, mvsc-ghgs}, the reconstruction performance is evaluated by using three metrics, i.e., Peak Signal-to-Noise Ratio (PSNR), Structural Similarity Index (SSIM)~\cite{ssim}, and Learned Perceptual Image Patch Similarity (LPIPS)~\cite{lpips}.
PSNR and SSIM measure pixel-wise similarity and structural fidelity between the rendered image and the ground truth, while LPIPS reflects the perceptual similarity in the feature domain.
We additionally report frames per second (FPS) to evaluate the inference speed.
Since FPS values of our method vary depending on the number of foreground tokens per input image, we report the average value measured on the THuman2.0 dataset~\cite{thuman}.

\begin{figure*}[t]
    \centering
    \includegraphics[width=0.85\textwidth]{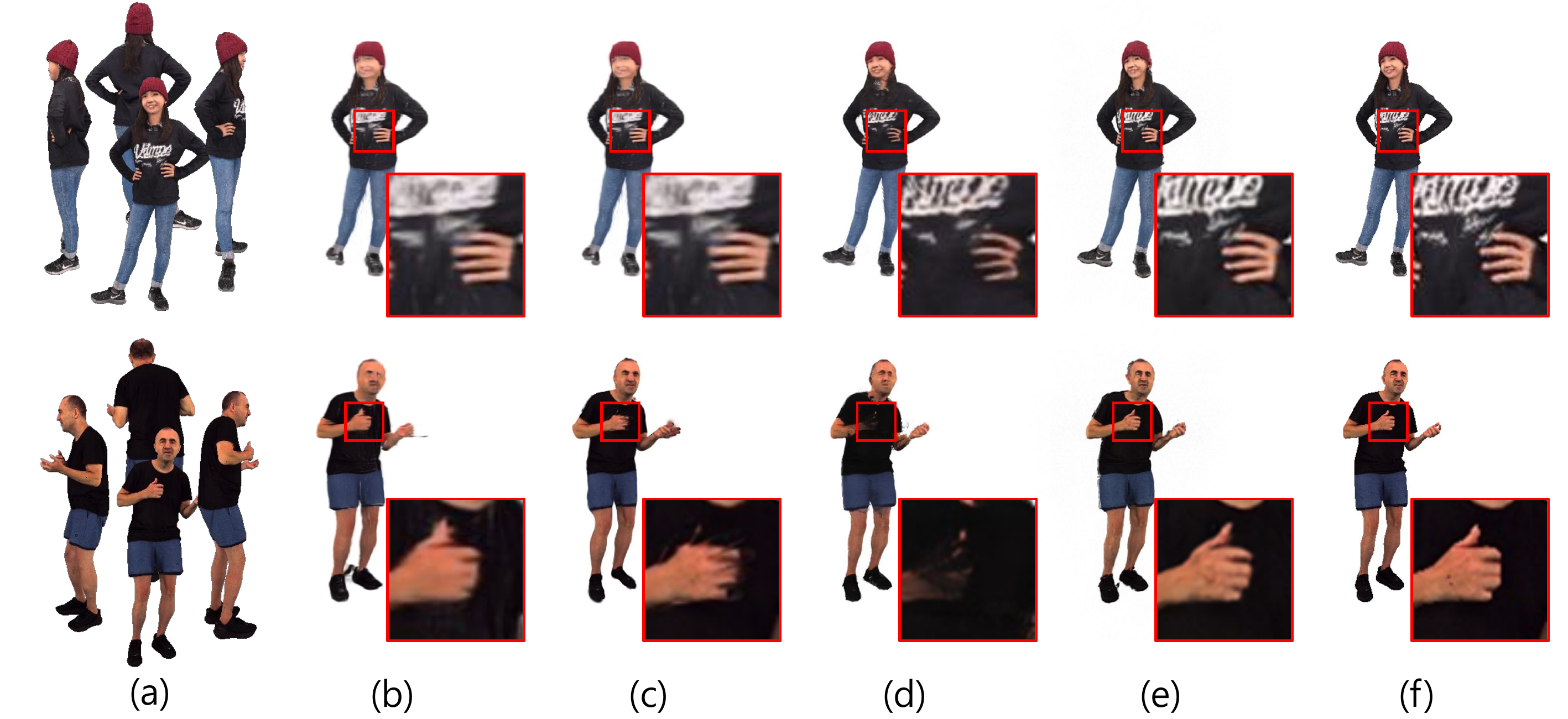}
    \caption{\textbf{Qualitative comparison of cross-domain novel view synthesis on 2K2K~\cite{2k2k} and X-Humans~\cite{xhuman} datasets  (from top to bottom).} (a) Input images. (b) Results by GPS-Gaussian~\cite{gps-gaussian}. (c) Results by GHG~\cite{ghg}. (d) Results by RoGSplat~\cite{rogsplat}. (e) Results by the proposed method. (f) Ground truth.}
    \label{fig:fig5}
\end{figure*}

\begin{table}[!t]
\centering
\caption{\textbf{Quantitative comparison of cross-domain novel view synthesis on 2K2K~\cite{2k2k} and X-Humans~\cite{xhuman} datasets.} The best result is shown in bold.}
\label{tab:cross_syn}
\footnotesize
\setlength{\tabcolsep}{4pt}
\begin{tabular}{lcccccc}
\toprule
\multirow{2}{*}{Methods} & \multicolumn{3}{c}{THuman2.0 $\to$ 2K2K} & \multicolumn{3}{c}{THuman2.0 $\to$ X-Humans}\\
\cmidrule(lr){2-4}\cmidrule(lr){5-7}
& PSNR$\uparrow$ & SSIM$\uparrow$ & LPIPS$\downarrow$ & PSNR$\uparrow$ & SSIM$\uparrow$ & LPIPS$\downarrow$ \\
\midrule
GPS-Gaussian~\cite{gps-gaussian} & 24.47 & 0.9361 & 0.0721 & 24.34 & 0.9330 & 0.0795 \\
GHG~\cite{ghg} & 26.74 & 0.9548 & 0.0518 & 26.06 & 0.9494 & 0.0580 \\
RoGSplat~\cite{rogsplat} & 26.96 & 0.9490 & 0.0489 & 26.57 & 0.9496 & 0.0507 \\
GHGS-MVSC~\cite{mvsc-ghgs} & 28.59 & 0.9527 & 0.0480 & 25.17 & 0.9375 & 0.0674 \\
\textbf{Ours} & \textbf{29.52} & \textbf{0.9580} & \textbf{0.0412} & \textbf{27.53} & \textbf{0.9499} & \textbf{0.0474} \\
\bottomrule
\end{tabular}
\end{table}

\subsection{Performance evaluation}
\noindent\textbf{Quantitative evaluation.} To demonstrate the effectiveness of the proposed method, we compare our method with previous generalizable human Gaussian splatting methods, i.e., GPS-Gaussian~\cite{gps-gaussian}, GHG~\cite{ghg}, RoGSplat~\cite{rogsplat}, and GHGS-MVSC~\cite{mvsc-ghgs}. 
Note that GPS-Gaussian is evaluated under a 6-view setting following~\cite{rogsplat}, while other methods use 4-view inputs.
The result of in-domain generalization comparisons is shown in Table~\ref{tab:in_domain_others}.
As can be seen, the proposed method consistently achieves the best performance across all three datasets and all three metrics. 
Specifically, the proposed method achieves 32.42 dB PSNR on THuman2.0, 31.68 dB on 2K2K, and 31.96 dB on X-Humans, which outperforms the second-best method by 1.49 dB, 1.01 dB, and 1.36 dB, respectively. 
Similar improvements are observed in SSIM and LPIPS metrics, which indicates that the proposed method better represents both pixel-wise and perceptual fidelity. 
Notably, the proposed method achieves these improvements while running at 33.96 FPS, which is substantially faster than RoGSplat (5.42 FPS) and GHGS-MVSC (2.79 FPS).
The cross-domain generalization comparison is also presented in Table~\ref{tab:cross_syn}. 
Our approach improves PSNR values by a notable margin, which demonstrates the robust generalization capability compared to previous methods (+0.93 dB over GHGS-MVSC on 2K2K and +0.96 dB over RoGSplat on X-Humans).

\noindent\textbf{Qualitative evaluation.} The qualitative comparisons on in-domain and cross-domain evaluations are presented in Figs.~\ref{fig:fig4} and~\ref{fig:fig5}, respectively.
Compared to existing methods, the proposed method reconstructs the target-view human image accurately in fine-grained regions, such as fingers, facial details, and textured clothes (see Fig.~\ref{fig:fig4}(e) and Fig.~\ref{fig:fig5}(e)).
In contrast, previous methods often fail to faithfully recover such intricate details due to the insufficient Gaussian density, which leads to the noticeable degradation in these regions (see Figs.~\ref{fig:fig4}(b)(c)(d) and Figs.~\ref{fig:fig5}(b)(c)(d)).
These visual comparisons confirm the effectiveness of the proposed method in faithfully reconstructing fine-scale human details.
\begin{figure*}[t]
    \centering
    \includegraphics[width=\textwidth]{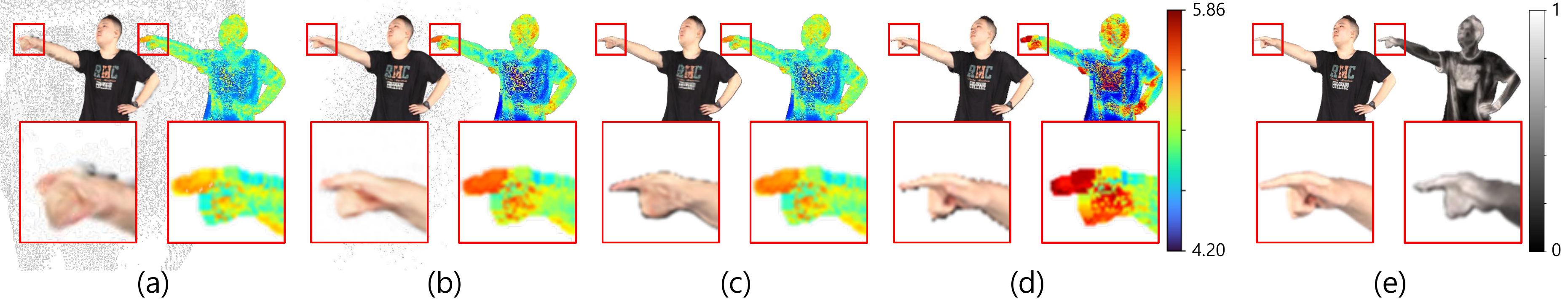}
    \caption{\textbf{Ablation studies on the THuman2.0 dataset.} We visualize the rendering result and the local Gaussian density. (a) Results by the baseline. (b) Results without patch supervision. (c) Results without $\mathcal{L}_{\text{comp}}$. (d) Results by the proposed method. (e) Ground truth.}
    \label{fig:fig6}
    \vspace{-0.7em}
    
\end{figure*}

\subsection{Ablation study}
\noindent\textbf{Effectiveness of the proposed components.}
In Table~\ref{tab:ablation}, we conduct ablation studies on the THuman2.0 dataset to validate the effectiveness of the proposed components. The complexity-guided 
\begin{wraptable}[9]{r}{0.37\textwidth}
\vspace{-1.3em}
\centering
\caption{\textbf{Ablation studies on the THuman2.0 dataset.}}
\label{tab:ablation}
\vspace{1.0em}
\footnotesize
\setlength{\tabcolsep}{2.0pt}
\begin{tabular}{ccccc}
\toprule
$\mathcal{L}_{\text{comp}}$ & Patch sup. & PSNR$\uparrow$ & SSIM$\uparrow$ & LPIPS$\downarrow$ \\
\midrule
\ding{55} & \ding{55} & 30.32 & 0.9664 & 0.0316 \\
\ding{51} & \ding{55} & 30.79 & 0.9695 & 0.0269 \\
\ding{55} & \ding{51} & 32.06 & 0.9752 & 0.0209 \\
\ding{51} & \ding{51} & \textbf{32.42} & \textbf{0.9766} & \textbf{0.0201} \\
\bottomrule
\end{tabular}
\end{wraptable}
regularization scheme improves PSNR by 0.47 dB and noticeably reduces the LPIPS value, which confirms that the explicit guidance of the Gaussian density by the local complexity is effective for reconstructing fine-grained details.
The complexity-aware patch-based supervision strategy substantially improves all metrics by concentrating the supervisory signal on intricate regions.
Consequently, the synergy among the proposed components encourages the local Gaussian density to be positively correlated with the complexity prior, which yields superior performance in generalizable human Gaussian splatting, as shown in Fig.~\ref{fig:fig6}.

\noindent\textbf{Performance variations according to different $k$ values.}
We investigate the influence of the number of nearest neighbors $k$ for computing the local density of Gaussians as defined in Eq.~\ref{eq:knn_density}. As 
\begin{wraptable}[8]{r}{0.47\textwidth}
\vspace{-1.3em}
\centering
\caption{\textbf{Performance analysis according to different $k$ values on the THuman2.0 dataset.}}
\label{tab:k_values}
\vspace{1.0em}
\footnotesize
\setlength{\tabcolsep}{7.5pt}
\begin{tabular}{lccc}
\toprule
\# Neighbors ($k$) & PSNR$\uparrow$ & SSIM$\uparrow$ & LPIPS$\downarrow$ \\
\midrule
$k=3$ & 30.55 & 0.9684 & 0.0299 \\
$k=5$ & 30.68 & 0.9690 & 0.0275 \\
$k=7$ & \textbf{30.79} & 0.9695 & \textbf{0.0269} \\
$k=9$ & 30.75 & \textbf{0.9696} & 0.0271 \\
\bottomrule
\end{tabular}
\end{wraptable}
shown in Table~\ref{tab:k_values}, the performance improves as $k$ increases from 3 to 7, where $k=7$ achieves the best PSNR and LPIPS values.
When $k$ is further increased to 9, the performance remains comparable across all metrics.
This suggests that an excessively small $k$ makes the density score sensitive to the noise.
Based on the trade-off between the performance and the computational cost, we set $k=7$ in our experiments.

\noindent\textbf{Effectiveness of complexity-aware patch sampling.}
For our patch-based supervision, patch centers are sampled with probabilities proportional to the local complexity. To validate this strategy,
\begin{wraptable}[8]{r}{0.45\textwidth}
\vspace{-1.3em}
\centering
\caption{\textbf{Comparison of patch sampling strategies on the THuman2.0 dataset.}}
\label{tab:patch_sampling}
\vspace{1.0em}
\footnotesize
\setlength{\tabcolsep}{1.4pt}
\begin{tabular}{lccc}
\toprule
Strategy & PSNR$\uparrow$ & SSIM$\uparrow$ & LPIPS$\downarrow$ \\
\midrule
Random sampling & 31.63 & 0.9731 & 0.0227 \\
Gradient-based sampling & 31.81 & 0.9744 & 0.0216 \\
\textbf{Complexity-aware sampling} & \textbf{32.06} & \textbf{0.9752} & \textbf{0.0209} \\
\bottomrule
\end{tabular}
\end{wraptable}
we compare the complexity-aware sampling with two alternative strategies, as shown in Table~\ref{tab:patch_sampling}.
For the image-gradient-based sampling, the gradient magnitude map computed by the Sobel operator with $3\times3$ kernels is used as the sampling probability instead of the 2D complexity map.
The complexity-aware sampling consistently outperforms both alternatives across all metrics.
Specifically, the image-gradient-based sampling improves PSNR by only 0.18 dB over random sampling, whereas the complexity-aware sampling achieves an improvement of 0.43 dB.
This confirms that incorporating the local complexity of the human subject into the training scheme effectively guides the model to learn fine-grained details.

\begin{wraptable}[9]{r}{0.54\textwidth}
\vspace{-0.5em}
\centering
\caption{\textbf{Ablation study of complexity components on the THuman2.0 dataset.}}
\label{tab:complexity_components}
\vspace{1.0em}
\footnotesize
\setlength{\tabcolsep}{1pt}
\begin{tabular}{lccc}
\toprule
Complexity component & PSNR$\uparrow$ & SSIM$\uparrow$ & LPIPS$\downarrow$ \\
\midrule
Geometric complexity ($\bar{H}^{\text{geo}}$) & 31.65 & 0.9727 & 0.0232 \\
Textural complexity ($\bar{H}^{\text{tex}}$) & 32.24 & 0.9752 & 0.0214 \\
Average operation ($\mathrm{mean}(\bar{H}^{\text{geo}},\bar{H}^{\text{tex}})$) & 32.31 & 0.9760 & 0.0205 \\
\textbf{Maximum operation ($\max(\bar{H}^{\text{geo}},\bar{H}^{\text{tex}})$)} & \textbf{32.42} & \textbf{0.9766} & \textbf{0.0201} \\
\bottomrule
\end{tabular}
\end{wraptable}
\noindent\textbf{Effectiveness of complexity components.}
To examine the contribution of the geometric and textural complexities in Eq.~\ref{eq:complexity_prior}, we conduct an ablation study of complexity components.
As shown in Table~\ref{tab:complexity_components}, combining geometric and textural complexity yields better reconstruction quality than using either component alone. 
Furthermore, the maximum operation achieves the best performance across all metrics, outperforming the average operation by 0.11 dB in PSNR.
This result indicates that the maximum operation effectively preserves a strong complexity cue when either geometric or textural variation is prominent.

\noindent\textbf{Correlation between complexity and density of Gaussians.}
To directly verify that the
\begin{wraptable}[9]{r}{0.46\textwidth}
\vspace{-1.3em}
\centering
\caption{\textbf{Spearman rank correlation between local complexity and density of Gaussians on the THuman2.0 dataset.}}
\label{tab:spearman}
\vspace{1.0em}
\footnotesize
\setlength{\tabcolsep}{6pt}
\begin{tabular}{cc}
\toprule
Complexity-guided reg. & Correlation coefficient$\uparrow$ \\
\midrule
\ding{55} & 0.1086 \\
\ding{51} & \textbf{0.4262} \\
\bottomrule
\end{tabular}
\end{wraptable}
proposed regularization scheme changes the distribution of Gaussians, we measure the Spearman rank correlation between the local complexity and the local density of estimated Gaussians on the THuman2.0 dataset.
As shown in Table~\ref{tab:spearman}, the correlation coefficient increases from 0.1086 to 0.4262 with the complexity-guided regularization scheme.
This result confirms that the distribution of Gaussians is closely aligned with the complexity prior.

\noindent\textbf{Performance comparison under different regularization settings.}
Since the complexity prior is computed from the textured clothed human mesh, the proposed method relies on an additional
\begin{wraptable}[12]{r}{0.62\textwidth}
\vspace{-1.3em}
\centering
\caption{\textbf{Quantitative comparison under different regularization settings on the THuman2.0 dataset.}}
\label{tab:regularization_settings}
\vspace{1.0em}
\footnotesize
\setlength{\tabcolsep}{2pt}
\begin{tabular*}{\linewidth}{@{\extracolsep{\fill}}lccccc@{}}
\toprule
Methods & Depth reg. & Textured mesh & PSNR$\uparrow$ & SSIM$\uparrow$ & LPIPS$\downarrow$ \\
\midrule
GPS-Gaussian~\cite{gps-gaussian} & \ding{51} & \ding{55} & 26.54 & 0.9539 & 0.0610 \\
RoGSplat~\cite{rogsplat} & \ding{51} & \ding{55} & 28.83 & 0.9615 & 0.0429 \\
GHGS-MVSC~\cite{mvsc-ghgs} & \ding{55} & \ding{55} & 30.93 & 0.9710 & 0.0334 \\
\midrule
\multirow{3}{*}{\textbf{Ours}} & \ding{55} & \ding{55} & 31.83 & 0.9737 & 0.0231 \\
 & \ding{51} & \ding{55} & 32.04 & 0.9744 & 0.0224 \\
 & \ding{51} & \ding{51} & \textbf{32.42} & \textbf{0.9766} & \textbf{0.0201} \\
\bottomrule
\end{tabular*}
\end{wraptable}
3D prior which is not used by previous methods.
For a fair comparison, we evaluate a variant of the proposed method which computes the complexity prior directly from input images by applying Eqs.~\ref{eq:geo_entropy} and~\ref{eq:tex_entropy} to normal maps predicted by an off-the-shelf monocular normal estimator~\cite{sapiens} and RGB pixels, respectively.
As shown in Table~\ref{tab:regularization_settings}, the proposed method consistently outperforms previous methods under both regularization settings.
Without the depth regularization scheme, the proposed method achieves 31.83 dB in PSNR, which surpasses GHGS-MVSC~\cite{mvsc-ghgs} by 0.90 dB.
When the depth regularization scheme is additionally employed, the proposed method reaches 32.04 dB in PSNR, which outperforms RoGSplat~\cite{rogsplat} by 3.21 dB.
These results are slightly lower than those obtained with a textured mesh, which may be attributed to inaccuracies in the image-derived complexity maps. This is because complexity maps derived from a monocular normal estimator may be inconsistent across views.
Nevertheless, the competitive performance of this variant suggests that the proposed method can be extended to settings where textured meshes are unavailable.
\section{Conclusion}
In this paper, we introduce a novel method for generalizable human Gaussian splatting.
The key idea of the proposed method is to allocate Gaussians in proportion to the local complexity of a clothed human, which is quantified through the entropy of geometric and textural variations.
This approach allocates Gaussians densely in fine-grained regions, which effectively contributes to the faithful reconstruction of intricate human details.
Additionally, the proposed complexity-aware patch-based supervision concentrates the supervisory signal on complex regions, which provides the model with sufficient supervision on fine-scale details.
As a result, the proposed method consistently produces high-fidelity rendering results.
Experimental results demonstrate the effectiveness of our approach in generalizable human Gaussian splatting on both in-domain and cross-domain settings.


\begin{ack}
This work was supported by the Institute of Information \& Communications Technology Planning \& Evaluation (IITP) grant funded by the Korea government through the Ministry of Science and ICT (MSIT) (No. 2018-0-00207, RS-2018-II180207, Immersive Media Research Laboratory).
\end{ack}
\bibliographystyle{unsrtnat}

\begingroup
\small
\bibliography{references}
\endgroup
\clearpage
\appendix
\section*{Appendix --- Complexity-guided Regularization for Generalizable Human Gaussian Splatting}

The appendix is structured as follows:
\begin{itemize}
    \item We describe further implementation details in Sec.~\ref{sec:impl_details}.
    \item We present additional ablation studies of the proposed method in Sec.~\ref{sec:more_ablation}.
    \item We provide additional comparisons with previous methods in Sec.~\ref{sec:more_comparisons}.
    \item We discuss limitations and future directions in Sec.~\ref{sec:limitations}.
    \item We provide the ethics statement in Sec.~\ref{sec:ethics}.
     \item We also submit a supplementary video which includes additional qualitative comparisons.
\end{itemize}
\section{Implementation details}
\label{sec:impl_details}

\subsection{Network architecture}
The tokenizer is implemented as a single linear layer, which embeds each non-overlapping $8\times8$ patch to a $1024$-dimensional token.
The transformer encoder consists of $24$ self-attention layers, where the hidden dimension and the number of heads are set to $1024$ and $64$, respectively.
The Gaussian decoder consists of a LayerNorm layer followed by a single linear layer.
The linear layer produces an output feature $\in\mathbb{R}^{K\times 15}$ for each token (where $K=64$), which corresponds to attributes of $K$ Gaussians, i.e., depth, 3D spatial offset, scale, rotation quaternion, opacity, and color.
The output dimension and the activation function of each attribute are summarized in Table~\ref{tab:gaussian_decoder}.

\begin{table}[h]
\centering
\caption{\textbf{The output dimension and the activation function of each Gaussian attribute.}}
\label{tab:gaussian_decoder}
\footnotesize
\setlength{\tabcolsep}{8pt}
\begin{tabular}{lccc}
\toprule
Attribute & Per-token Dim. & Per-Gaussian Dim. & Activation \\
\midrule
Depth                  & 64  & 1 & Sigmoid \\
3D spatial offset      & 192 & 3 & - \\
Scale                  & 192 & 3 & Exp \\
Rotation               & 256 & 4 & Norm. \\
Opacity                & 64  & 1 & Sigmoid \\
Color                  & 192 & 3 & - \\
\midrule
Total        & 960 & 15 & \\
\bottomrule
\end{tabular}
\end{table}

\subsection{Training details}
The proposed model is trained for $800{,}000$ iterations with a batch size of $2$.
Training takes approximately $3.5$ days on a single NVIDIA RTX 4090 GPU.
We use the AdamW optimizer, where the weight decay coefficient is set to $0.05$ for all parameters except biases and normalization parameters.
The learning rate is linearly warmed up over the first $2{,}000$ iterations, and decayed to $0$ via the cosine schedule function afterwards.
We apply gradient clipping throughout training, where the maximum norm is set to $0.5$.
The complexity prior $\{c_i\}$ is precomputed before training to avoid the redundant computation.
The small constant $\epsilon$ in Eq.~\ref{eq:knn_density} is set to $1\times10^{-6}$ for the numerical stability.

\section{Additional ablation studies}
\label{sec:more_ablation}

\subsection{Effectiveness of the proposed components without depth regularization scheme}
While the effectiveness of the proposed components is checked with the depth regularization term $\mathcal{L}_{\text{depth}}$ in the manuscript, we further verify their effectiveness when $\mathcal{L}_{\text{depth}}$ is not applied.
As shown in Table~\ref{tab:no_depth_ablation}, the complexity-guided regularization scheme $\mathcal{L}_{\text{comp}}$ improves PSNR by $0.61$ dB, while the patch-based supervision strategy improves PSNR by $1.16$ dB.
Consequently, the synergy between the two components yields a total improvement of $2.21$ dB in PSNR over the baseline.
This confirms that the effectiveness of the proposed components is preserved regardless of the presence of $\mathcal{L}_{\text{depth}}$.

\begin{table}[h]
\centering
\caption{\textbf{Ablation studies on the THuman2.0 dataset without the depth regularization scheme $\mathcal{L}_{\text{depth}}$.}}
\label{tab:no_depth_ablation}
\footnotesize
\setlength{\tabcolsep}{6pt}
\begin{tabular}{ccccc}
\toprule
$\mathcal{L}_{\text{comp}}$ & Patch sup. & PSNR$\uparrow$ & SSIM$\uparrow$ & LPIPS$\downarrow$ \\
\midrule
\ding{55} & \ding{55} & 29.80 & 0.9632 & 0.0357 \\
\ding{51} & \ding{55} & 30.41 & 0.9665 & 0.0327 \\
\ding{55} & \ding{51} & 30.96 & 0.9702 & 0.0279 \\
\ding{51} & \ding{51} & \textbf{32.01} & \textbf{0.9750} & \textbf{0.0212} \\
\bottomrule
\end{tabular}
\end{table}

\subsection{Effectiveness of the pairwise monotonic ranking formulation}
The complexity-guided regularization scheme $\mathcal{L}_{\text{comp}}$ in the manuscript adopts a pairwise monotonic ranking formulation in Eq.~\eqref{eq:rank_loss}.
We analyze the effectiveness of our formulation by comparing it with a Pearson correlation-based formulation~\cite{pearson}.
This formulation maximizes the linear correlation between complexity values and local Gaussian density scores across the $M$ sampled surface points as follows:
\begin{equation}
\mathcal{L}_{\text{pearson}} = 1 - \frac{\sum_{m=1}^{M} (c_m - \bar{c})(s_m - \bar{s})}{\sqrt{\sum_{m=1}^{M} (c_m - \bar{c})^2} \cdot \sqrt{\sum_{m=1}^{M} (s_m - \bar{s})^2}},
\label{eq:lpearson}
\end{equation}
where $c_m$ and $s_m$ denote the complexity value and the local Gaussian density score at the $m$-th sampled surface point, respectively. $\bar{c}$ and $\bar{s}$ denote the average of  complexity values and local Gaussian density scores across $M$ samples, respectively.
As shown in Table~\ref{tab:loss_formulation}, the Pearson correlation-based formulation yields a $0.0015$ drop in SSIM value and a $0.0009$ increase in LPIPS value compared to the baseline which is trained without the complexity-guided regularization scheme.
This suggests that the Pearson correlation-based formulation introduces an unnecessary linearity assumption.
In particular, while Gaussians should be placed densely in complex regions, the density score is not required to be linearly proportional to the complexity value.
In contrast, our pairwise monotonic ranking formulation improves PSNR by $0.47$ dB and the SSIM value by $0.0031$ compared to the baseline.
These results confirm the effectiveness of our pairwise monotonic ranking formulation for complexity-guided regularization, as it only encourages density scores to preserve the order induced by complexity values.

\begin{table}[h]
\centering
\caption{\textbf{Comparison of loss formulations for the complexity-guided regularization scheme on the THuman2.0 dataset.}}
\label{tab:loss_formulation}
\footnotesize
\setlength{\tabcolsep}{8pt}
\begin{tabular}{lccc}
\toprule
Loss formulation & PSNR$\uparrow$ & SSIM$\uparrow$ & LPIPS$\downarrow$ \\
\midrule
Baseline (w/o complexity-guided regularization scheme)        & 30.32 & 0.9664 & 0.0316 \\
Pearson correlation formulation $\mathcal{L}_{\text{pearson}}$            & 30.31 & 0.9649 & 0.0325 \\
\textbf{Pairwise ranking formulation Eq.~\eqref{eq:rank_loss}}  & \textbf{30.79} & \textbf{0.9695} & \textbf{0.0269} \\
\bottomrule
\end{tabular}
\end{table}

\subsection{Robustness to perturbations at training and inference phases}
We examine the robustness of the proposed method to perturbations at training and inference time on the 2K2K dataset.
For the training-time perturbation, we reduce the number of mesh vertices to 50\% and add Gaussian noise with $\sigma=0.02$ to the UV texture and geometry of the human mesh, which
\begin{wraptable}[10]{r}{0.5\textwidth}
\vspace{-1.0em}
\centering
\caption{\textbf{Robustness to degraded training-time regularization and inference-time mask perturbation on the 2K2K dataset.}}
\label{tab:robustness}
\vspace{1.0em}
\footnotesize
\setlength{\tabcolsep}{4pt}
\begin{tabular}{cccc}
\toprule
Perturbed regularization & Input masks & PSNR$\uparrow$ & $\Delta$PSNR \\
\midrule
\ding{55} & Clean  & 31.68 & -- \\
\ding{55} & Jitter & 31.41 & $-0.27$ \\
\ding{51} & Clean  & 31.46 & $-0.22$ \\
\ding{51} & Jitter & 31.28 & $-0.40$ \\
\bottomrule
\end{tabular}
\end{wraptable}
degrades the regularization signal during the training phase.
For the inference-time perturbation, we apply jitter noise with $\sigma=4$ pixels to the input foreground masks.
As shown in Table~\ref{tab:robustness}, the inference-time and training-time perturbations reduce PSNR by 0.27 dB and 0.22 dB, respectively.
When both perturbations are applied simultaneously, the performance drop remains limited to 0.40 dB in PSNR, which indicates that the proposed method is robust to moderate errors at both training and test time.

\section{Additional comparisons}
\label{sec:more_comparisons}

\subsection{Analysis of the computational efficiency}
We compare the number of parameters and the GPU memory usage of the proposed method with previous generalizable human Gaussian splatting methods.
As shown in Table~\ref{tab:cost}, GHGS-MVSC~\cite{mvsc-ghgs}, which incorporates a foundation model~\cite{vggt} as its backbone, contains the largest number of parameters ($987.2$~M) and requires the highest GPU memory ($10.47$~GB).
Although RoGSplat~\cite{rogsplat} contains only $12.3$~M parameters, it requires $4.21$~GB of GPU memory due to the volumetric feature representation, which stores feature vectors across 3D voxel grids.
In contrast, the proposed method contains $303.5$~M parameters and requires only $2.06$~GB of GPU memory, which achieves lower memory usage than that of RoGSplat~\cite{rogsplat}.
This is mainly because the proposed method discards tokens whose corresponding patches lie entirely outside the human mask, which substantially reduces the memory usage.
Furthermore, since the number of foreground tokens varies according to input images, the inference speed of the proposed method also varies during the inference.
As shown in Table~\ref{tab:fps_variation}, the proposed method runs at over $26$ FPS even in the worst case with $3{,}056$ foreground tokens, and reaches up to $45.65$ FPS in the best case with $1{,}437$ foreground tokens.
This demonstrates that the proposed method consistently maintains the real-time inference speed under various conditions.

\vspace{4pt}
\begin{table}[h]
\begin{minipage}[t]{0.6\textwidth}
\centering
\caption{\textbf{Comparison of model size and memory on the THuman2.0 dataset.}}
\label{tab:cost}
\footnotesize
\setlength{\tabcolsep}{6pt}
\begin{tabular}{lcc}
\toprule
Methods & \# of parameters (M) & GPU Memory (GB) \\
\midrule
GPS-Gaussian~\cite{gps-gaussian} & 5.1 & 0.46 \\
RoGSplat~\cite{rogsplat} & 12.3 & 4.21 \\
GHGS-MVSC~\cite{mvsc-ghgs} & 987.2 & 10.47 \\
\textbf{Ours} & 303.5 & 2.06 \\
\bottomrule
\end{tabular}
\end{minipage}
\hfill
\begin{minipage}[t]{0.4\textwidth}
\centering
\caption{\textbf{FPS variations on the THuman2.0 dataset.}}
\label{tab:fps_variation}
\footnotesize
\setlength{\tabcolsep}{8pt}
\begin{tabular}{lccc}
\toprule
 & Worst & Mean & Best \\
\midrule
\# of tokens     & 3{,}056 & 2{,}258 & 1{,}437 \\
FPS $\uparrow$ & 26.35   & 33.96   & 45.65 \\
\bottomrule
\end{tabular}
\end{minipage}
\end{table}

\subsection{Comparison with different numbers of input views}
\begin{figure*}[h]
    \centering
    \includegraphics[width=\textwidth]{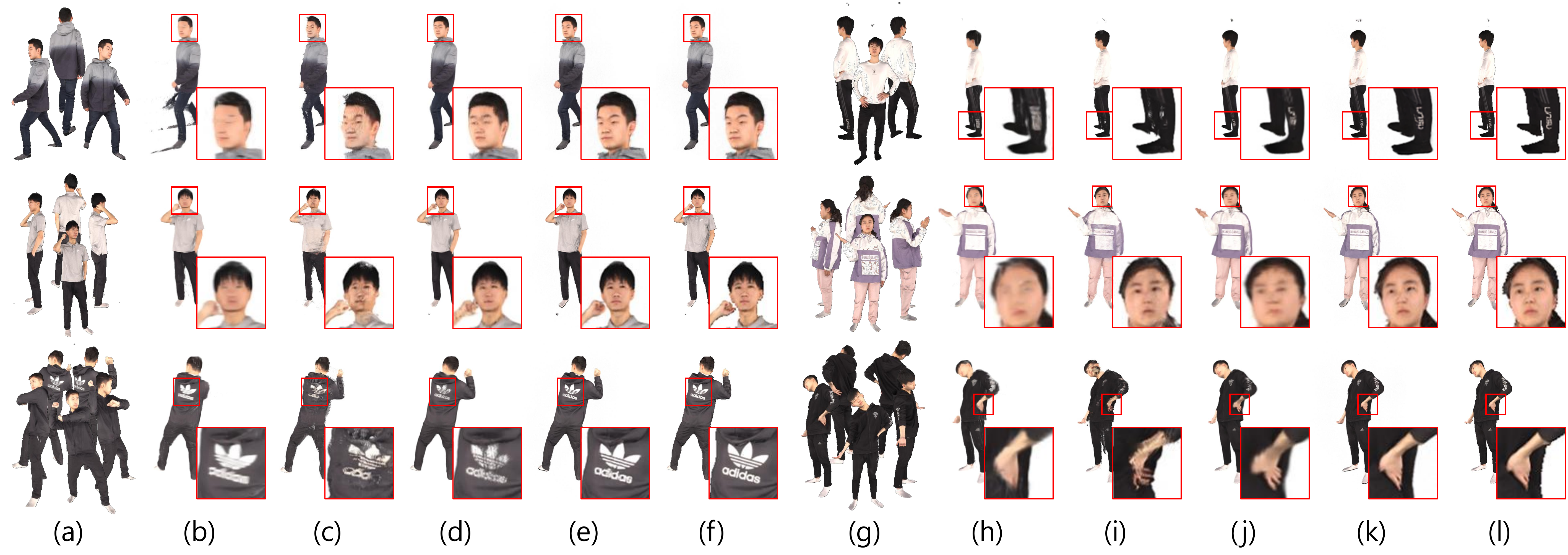}
    \caption{\textbf{Qualitative comparison of in-domain novel view synthesis on THuman2.0~\cite{thuman} with different numbers of input views (3-view, 4-view, and 5-view settings from top to bottom).} (a, g) Input images. (b, h) Results by GPS-Gaussian~\cite{gps-gaussian}. (c, i) Results by RoGSplat~\cite{rogsplat}. (d, j) Results by GHGS-MVSC~\cite{mvsc-ghgs}. (e, k) Results by the proposed method. (f, l) Ground truth.}

    \label{fig:input_views}
\end{figure*}

We further compare the proposed method with previous generalizable human Gaussian splatting methods under different numbers of input views.
Specifically, we evaluate the proposed method on the THuman2.0 dataset~\cite{thuman} under three different input view settings, i.e., 3-view, 4-view, and 5-view.
Note that GPS-Gaussian~\cite{gps-gaussian} is evaluated under the 6-view setting following~\cite{rogsplat}, while EVA-Gaussian~\cite{eva-gaussian} is evaluated only at the 4-view and 5-view settings following~\cite{mvsc-ghgs}.
As shown in Table~\ref{tab:input_views}, the proposed method consistently achieves the best performance across all three settings and all three metrics.
Specifically, the performance gap between the proposed method and the second-best method (i.e., GHGS-MVSC~\cite{mvsc-ghgs}) reaches $2.43$~dB, $1.49$~dB, and $2.24$~dB in PSNR at the 3-view, 4-view, and 5-view settings, respectively.
This demonstrates that the proposed method consistently achieves the superior rendering quality regardless of the number of input views.
The qualitative comparison is presented in Fig.~\ref{fig:input_views}, which shows that the proposed method reconstructs fine-grained human details more faithfully than previous methods for all input view settings.

\begin{table}[h]
\centering
\caption{\textbf{Quantitative comparison of in-domain novel view synthesis on THuman2.0~\cite{thuman} with different numbers of input views.} The best result is shown in bold.}
\label{tab:input_views}
\footnotesize
\setlength{\tabcolsep}{4pt}
\begin{tabular}{lccccccccc}
\toprule
\multirow{2}{*}{Methods} & \multicolumn{3}{c}{3-view} & \multicolumn{3}{c}{4-view} & \multicolumn{3}{c}{5-view}\\
\cmidrule(lr){2-4}\cmidrule(lr){5-7}\cmidrule(lr){8-10}
& PSNR$\uparrow$ & SSIM$\uparrow$ & LPIPS$\downarrow$ & PSNR$\uparrow$ & SSIM$\uparrow$ & LPIPS$\downarrow$ & PSNR$\uparrow$ & SSIM$\uparrow$ & LPIPS$\downarrow$ \\
\midrule
GPS-Gaussian~\cite{gps-gaussian} & - & - & - & - & - & - & 26.54 & 0.9539 & 0.0610 \\
EVA-Gaussian~\cite{eva-gaussian} & - & - & - & 26.31 & 0.9555 & 0.0391 & 27.54 & 0.9614 & 0.0297 \\
GHG~\cite{ghg} & 24.84 & 0.9232 & 0.0719 & 25.84 & 0.9462 & 0.0633 & 26.53 & 0.9571 & 0.0398 \\
RoGSplat~\cite{rogsplat} & 26.32 & 0.9478 & 0.0530 & 28.83 & 0.9615 & 0.0429 & 30.98 & 0.9711 & 0.0341 \\
GHGS-MVSC~\cite{mvsc-ghgs} & 27.81 & 0.9521 & 0.0517 & 30.93 & 0.9710 & 0.0334 & 31.54 & 0.9729 & 0.0269 \\
\textbf{Ours} & \textbf{30.24} & \textbf{0.9673} & \textbf{0.0276} & \textbf{32.42} & \textbf{0.9766} & \textbf{0.0201} & \textbf{33.78} & \textbf{0.9818} & \textbf{0.0165} \\
\bottomrule
\end{tabular}
\end{table}

\subsection{Comparison on the THuman2.1 dataset}
\begin{figure*}[p]
    \centering
    \includegraphics[width=0.85\textwidth]{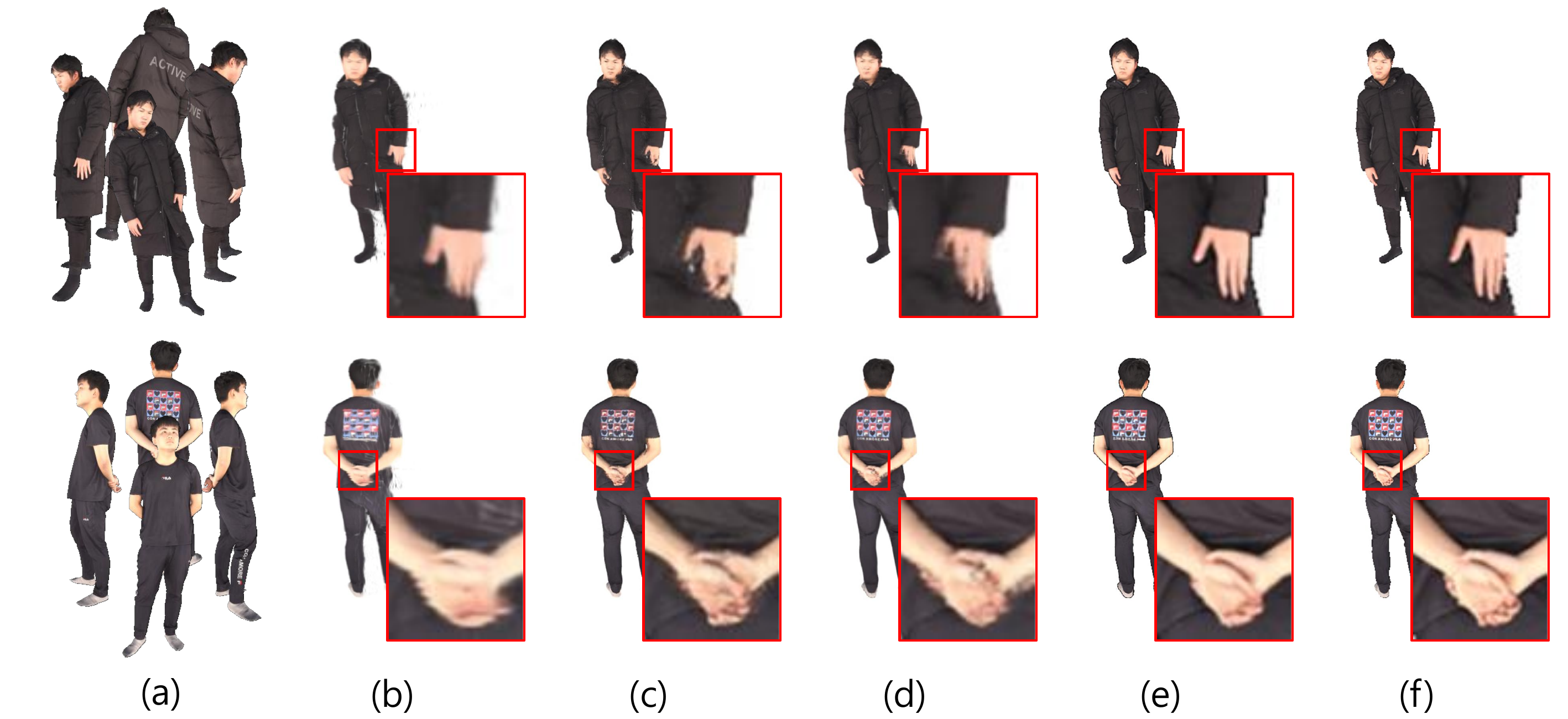}
    \caption{\textbf{Qualitative comparison of in-domain novel view synthesis on the THuman2.1~\cite{thuman} dataset.} (a) Input images. (b) Results by GPS-Gaussian~\cite{gps-gaussian}. (c) Results by RoGSplat~\cite{rogsplat}. (d) Results by GHGS-MVSC~\cite{mvsc-ghgs}. (e) Results by the proposed method. (f) Ground truth.}
    \label{fig:thuman21_in_domain}
\end{figure*}

\begin{figure*}[p]
    \centering
    \includegraphics[width=0.85\textwidth]{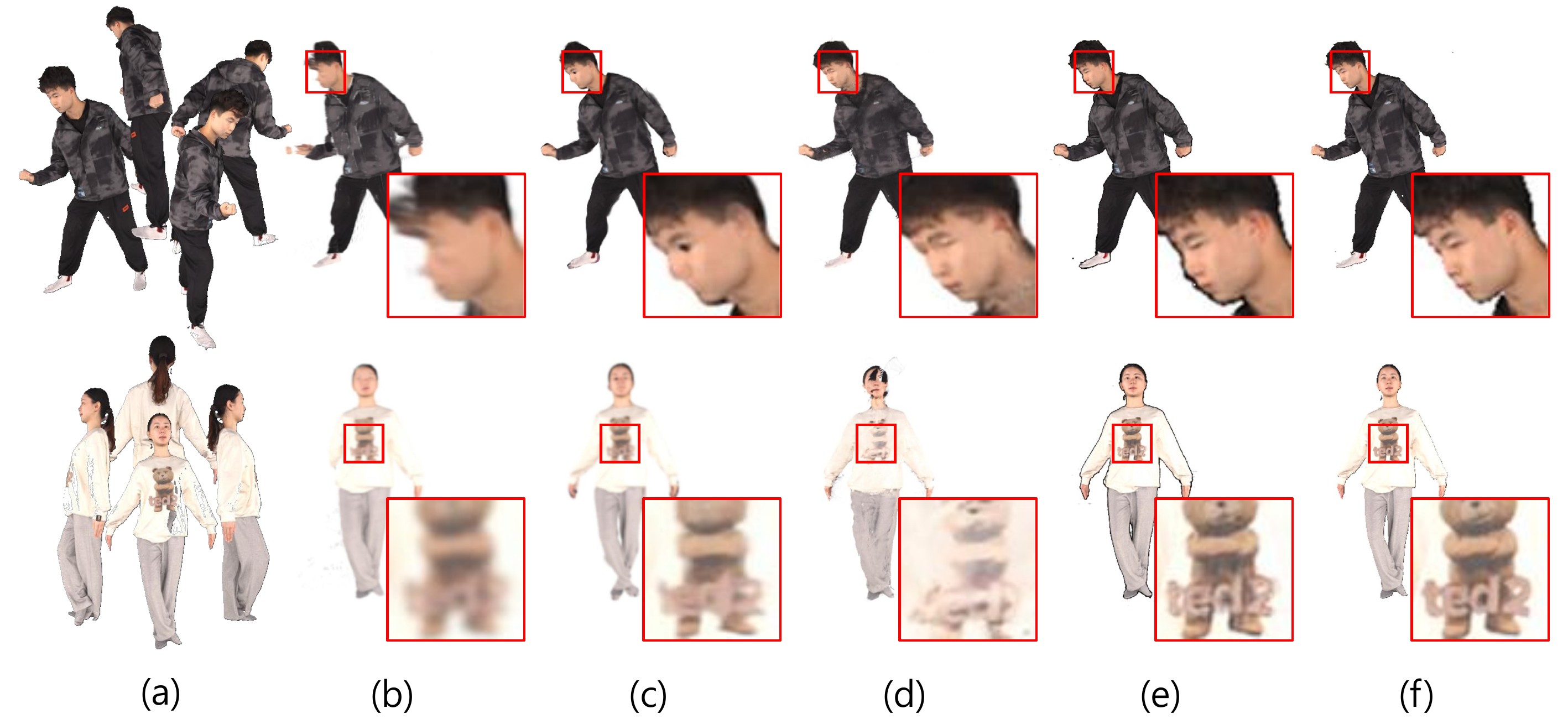}
    \caption{\textbf{Qualitative comparison of cross-domain novel view synthesis on the THuman2.1~\cite{thuman} dataset.} (a) Input images. (b) Results by GPS-Gaussian~\cite{gps-gaussian}. (c) Results by GHG~\cite{ghg}. (d) Results by RoGSplat~\cite{rogsplat}. (e) Results by the proposed method. (f) Ground truth.}
    \label{fig:thuman21_cross}
\end{figure*}

We further evaluate the proposed method on the THuman2.1~\cite{thuman} dataset under both in-domain and cross-domain settings.
For the in-domain setting, we follow training and test splits adopted in HART~\cite{hart}.
For the cross-domain setting, the proposed method and previous methods are trained on the THuman2.0 dataset~\cite{thuman} and directly evaluated on the THuman2.1 dataset without any fine-tuning.
As shown in Table~\ref{tab:thuman21}, the proposed method consistently achieves the best performance across all three metrics under both settings.
Specifically, the proposed method outperforms the second-best method (i.e., GHGS-MVSC~\cite{mvsc-ghgs}) by $4.38$~dB in the in-domain setting and by $1.26$~dB in the cross-domain setting.
This demonstrates that the proposed method consistently produces high-quality rendering results on the THuman2.1 dataset under both in-domain and cross-domain settings.
The qualitative comparisons under in-domain and cross-domain settings are also presented in Figs.~\ref{fig:thuman21_in_domain} and~\ref{fig:thuman21_cross}, respectively.

\begin{table}[h]
\centering
\caption{\textbf{Quantitative comparison of novel view synthesis on the THuman2.1~\cite{thuman} dataset under in-domain and cross-domain settings.} The best result is shown in bold.}
\label{tab:thuman21}
\footnotesize
\setlength{\tabcolsep}{6pt}
\begin{tabular}{lcccccc}
\toprule
\multirow{2}{*}{Methods} & \multicolumn{3}{c}{In-domain} & \multicolumn{3}{c}{THuman2.0 $\to$ THuman2.1} \\
\cmidrule(lr){2-4}\cmidrule(lr){5-7}
& PSNR$\uparrow$ & SSIM$\uparrow$ & LPIPS$\downarrow$ & PSNR$\uparrow$ & SSIM$\uparrow$ & LPIPS$\downarrow$ \\
\midrule
GPS-Gaussian~\cite{gps-gaussian} & 25.42 & 0.9430 & 0.0705 & 23.44 & 0.9315 & 0.0793 \\
GHG~\cite{ghg}                   & 27.83 & 0.9628 & 0.0498 & 25.07 & 0.9418 & 0.0651 \\
RoGSplat~\cite{rogsplat}         & 30.57 & 0.9717 & 0.0300 & 26.64 & 0.9524 & 0.0523 \\
GHGS-MVSC~\cite{mvsc-ghgs}       & 31.11 & 0.9735 & 0.0297 & 28.39 & 0.9621 & 0.0384 \\
\textbf{Ours}                    & \textbf{35.49} & \textbf{0.9851} & \textbf{0.0132} & \textbf{29.65} & \textbf{0.9623} & \textbf{0.0267} \\
\bottomrule
\end{tabular}
\end{table}

\begin{figure*}[p]
    \centering
    \includegraphics[width=0.85\textwidth]{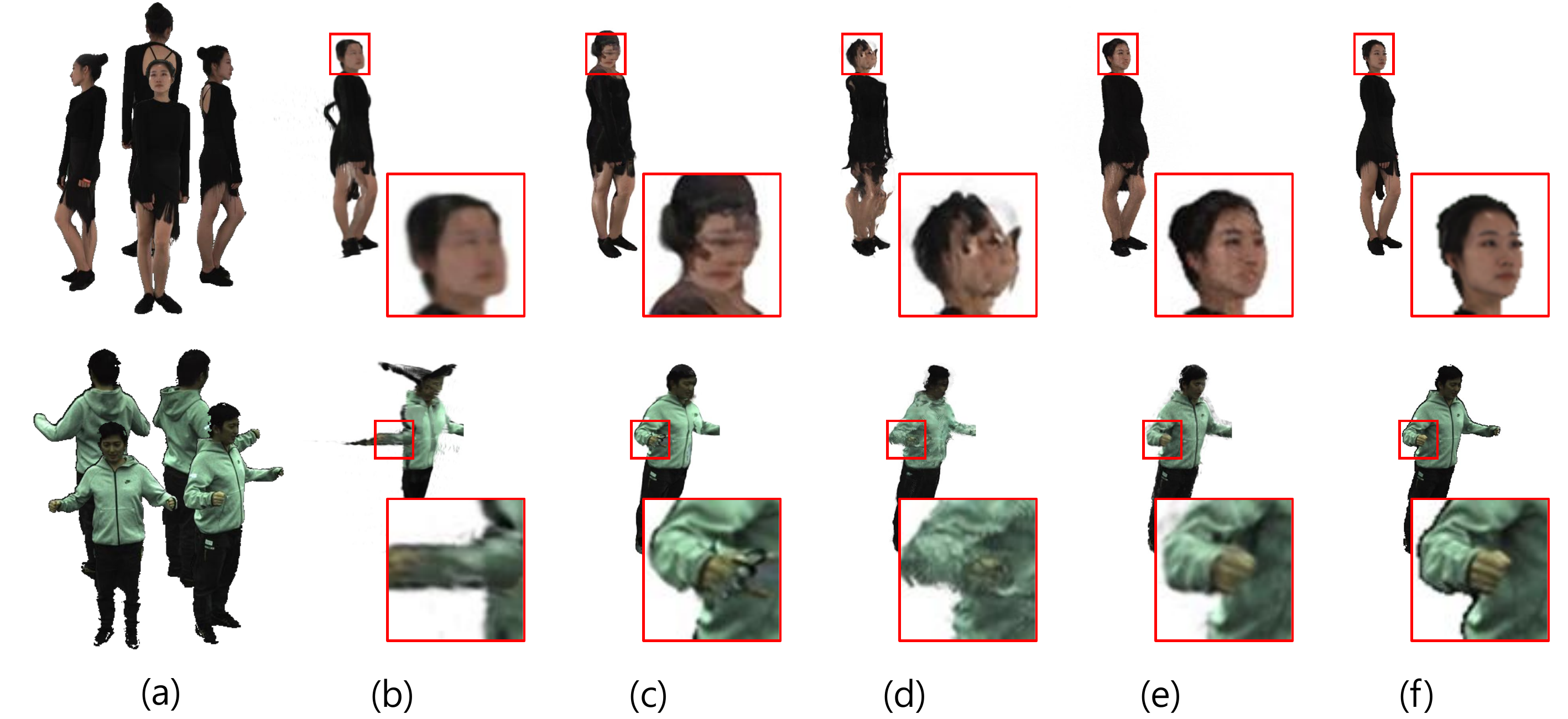}
    \caption{\textbf{Qualitative comparison of cross-domain novel view synthesis on DNA-Rendering~\cite{dna} and ZJU-MoCap~\cite{zju} datasets (from top to bottom).}  (a) Input images. (b) Results by GPS-Gaussian~\cite{gps-gaussian}. (c) Results by GHG~\cite{ghg}. (d) Results by RoGSplat~\cite{rogsplat}. (e) Results by the proposed method. (f) Ground truth.}
    \label{fig:real_world}
\end{figure*}

\subsection{Comparison on real-world datasets}
Beyond synthetic datasets, we additionally report the performance of the proposed method on real-world datasets, i.e., DNA-Rendering~\cite{dna} and ZJU-MoCap~\cite{zju}.
For the DNA-Rendering dataset, we follow training and test splits adopted in HART~\cite{hart}, while we use the official split for the ZJU-MoCap dataset.
The proposed method and previous methods are trained on the THuman2.0 dataset~\cite{thuman} and directly evaluated on the real-world datasets without any fine-tuning.
As shown in Table~\ref{tab:real_world}, the proposed method consistently achieves the best performance across all three metrics on both real-world datasets.
Specifically, the proposed method outperforms the second-best method (i.e., RoGSplat~\cite{rogsplat}) by $0.79$~dB in PSNR on DNA-Rendering~\cite{dna} and by $1.33$~dB in PSNR on ZJU-MoCap~\cite{zju}, while consistently improving the SSIM value and reducing the LPIPS value.
This result demonstrates the strong generalization capability of the proposed method to real-world human subjects.
The qualitative comparison is presented in Fig.~\ref{fig:real_world}.
On the DNA-Rendering dataset~\cite{dna} (see top row in Fig.~\ref{fig:real_world}), previous methods produce noticeably blurry facial features or visible artifacts around the face, whereas the proposed method clearly reconstructs the facial details.
Similarly, on the ZJU-MoCap dataset~\cite{zju} (see bottom row in Fig.~\ref{fig:real_world}), previous methods fail to faithfully recover the hand region, where fingers are geometrically distorted, while the proposed method preserves the structure of the fingers more accurately.
These visual comparisons confirm that the proposed method effectively reconstructs fine-grained human details even on real-world data.

\begin{table}[h]
\centering
\vspace{3px}
\caption{\textbf{Quantitative comparison of cross-domain novel view synthesis on real-world datasets, i.e., DNA-Rendering~\cite{dna} and ZJU-MoCap~\cite{zju}.}  The best result is shown in bold.}
\label{tab:real_world}
\footnotesize
\setlength{\tabcolsep}{6pt}
\begin{tabular}{lcccccc}
\toprule
\multirow{2}{*}{Methods} & \multicolumn{3}{c}{THuman2.0 $\to$ DNA-Rendering} & \multicolumn{3}{c}{THuman2.0 $\to$ ZJU-MoCap} \\
\cmidrule(lr){2-4}\cmidrule(lr){5-7}
& PSNR$\uparrow$ & SSIM$\uparrow$ & LPIPS$\downarrow$ & PSNR$\uparrow$ & SSIM$\uparrow$ & LPIPS$\downarrow$ \\
\midrule
GPS-Gaussian~\cite{gps-gaussian} & 23.22 & 0.8821 & 0.1296 & 26.26 & 0.9296 & 0.0829 \\
GHG~\cite{ghg}                   & 23.31 & 0.8852 & 0.0937 & 26.72 & 0.9357 & 0.0712 \\
RoGSplat~\cite{rogsplat}         & 25.68 & 0.9180 & 0.0803 & 27.66 & 0.9508 & 0.0547 \\
GHGS-MVSC~\cite{mvsc-ghgs}       & 17.76 & 0.8274 & 0.1397 & 23.10 & 0.8924 & 0.1056 \\
\textbf{Ours}                    & \textbf{26.47} & \textbf{0.9186} & \textbf{0.0788} & \textbf{28.99} & \textbf{0.9525} & \textbf{0.0536} \\
\bottomrule
\end{tabular}
\vspace{3px}
\end{table}

\subsection{Identity preservation on the face region}
A common concern in the human reconstruction task, as also raised in NoPo-Avatar~\cite{nopoavatar}, is the identity preservation, since the reconstructed subjects may suffer from identity shifts which hinder real-world applications.
To address this concern, we further evaluate the proposed method on the face region, which serves as the primary cue for the identity perception.
Specifically, we adopt the face quality evaluation protocol of NoPo-Avatar~\cite{nopoavatar} and GOHA~\cite{goha} on THuman2.0~\cite{thuman} and X-Humans~\cite{xhuman} datasets under the in-domain setting.

For each ground truth image, we first detect the face region using the pre-trained RetinaFace~\cite{retinaface}.
The detected bounding box is shared across all methods for the fair comparison.
On each face region, we measure PSNR, SSIM, and LPIPS to evaluate the pixel-wise similarity, the structural fidelity, and the perceptual similarity between the rendered face and the ground truth, respectively.
In addition, we adopt cosine similarity of identity embedding (CSIM)~\cite{arcface} and average keypoint distance (AKD)~\cite{fomm} as face-specific metrics. 
CSIM computes the cosine similarity between the rendered face and the ground truth in the feature space of ArcFace~\cite{arcface}, which is pre-trained to produce discriminative identity embeddings for face recognition.
AKD measures the geometric accuracy of facial structures by computing the average pixel distance between facial landmarks (e.g., eyes, nose, and mouth) detected from the rendered face and the ground truth.

\begin{figure}[h]
    \centering
    \includegraphics[width=1.0\textwidth]{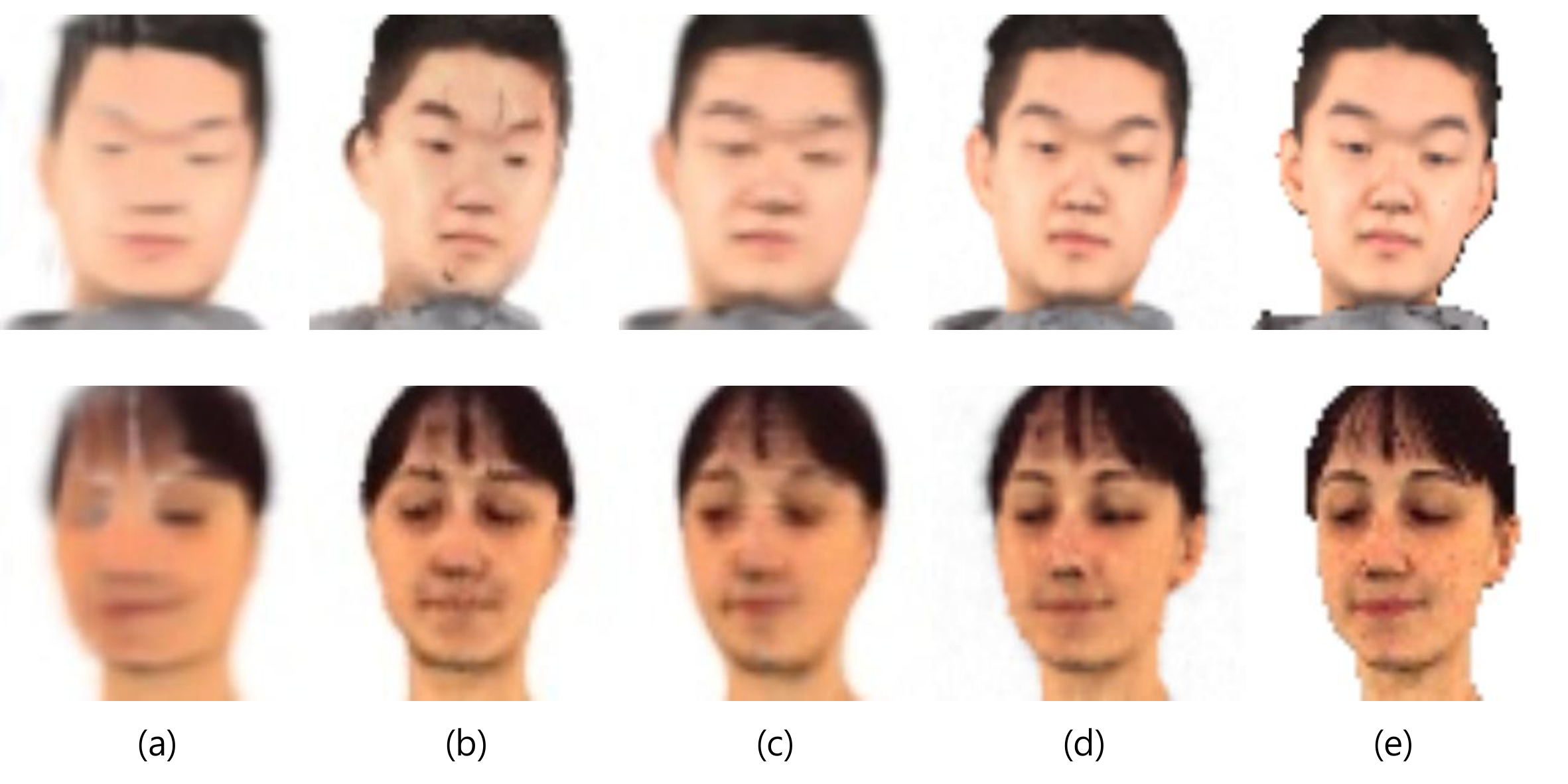}
    \caption{\textbf{Qualitative comparison on the face region of THuman2.0~\cite{thuman} and X-Humans~\cite{xhuman} datasets (from top to bottom).} (a) Results by GPS-Gaussian~\cite{gps-gaussian}. (b) Results by RoGSplat~\cite{rogsplat}. (c) Results by GHGS-MVSC~\cite{mvsc-ghgs}. (d) Results by the proposed method. (e) Ground truth.}
    \label{fig:face}
    \vspace{3px}
\end{figure}

As shown in Table~\ref{tab:face}, the proposed method consistently achieves the best performance across all five metrics on both datasets.
Specifically, the proposed method outperforms the second-best method (i.e., GHGS-MVSC~\cite{mvsc-ghgs}) by $1.50$~dB and $2.30$~dB in PSNR on the THuman2.0 and X-Humans datasets, respectively.
The proposed method also achieves the best performance in CSIM and AKD metrics on both datasets, which indicates that face regions of the rendered result faithfully preserve both the identity and the geometric structure.
The qualitative comparison is presented in Fig.~\ref{fig:face}, where the proposed method faithfully reconstructs facial features with the clear geometric structure.

\begin{table}[h]
\vspace{3px}
\centering
\caption{\textbf{Quantitative comparison on the face region of THuman2.0~\cite{thuman} and X-Humans~\cite{xhuman} datasets under in-domain settings.} The best result is shown in bold.}
\label{tab:face}
\footnotesize
\setlength{\tabcolsep}{4pt}
\begin{tabular}{lcccccccccc}
\toprule
\multirow{2}{*}{Methods} & \multicolumn{5}{c}{THuman2.0} & \multicolumn{5}{c}{X-Humans} \\
\cmidrule(lr){2-6}\cmidrule(lr){7-11}
& PSNR$\uparrow$ & SSIM$\uparrow$ & LPIPS$\downarrow$ & CSIM$\uparrow$ & AKD$\downarrow$
& PSNR$\uparrow$ & SSIM$\uparrow$ & LPIPS$\downarrow$ & CSIM$\uparrow$ & AKD$\downarrow$ \\
\midrule
GPS-Gaussian~\cite{gps-gaussian} & 20.20 & 0.798 & 0.240 & 0.426 & 3.19 & 19.52 & 0.712 & 0.327 & 0.388 & 3.69 \\
RoGSplat~\cite{rogsplat}         & 18.18 & 0.748 & 0.259 & 0.473 & 2.76 & 19.84 & 0.703 & 0.228 & 0.484 & 2.99 \\
GHGS-MVSC~\cite{mvsc-ghgs}       & 23.23 & 0.856 & 0.184 & 0.549 & 2.25 & 23.18 & 0.785 & 0.219 & 0.513 & 2.72 \\
\textbf{Ours}                    & \textbf{24.73} & \textbf{0.888} & \textbf{0.111} & \textbf{0.699} & \textbf{1.26} & \textbf{25.48} & \textbf{0.831} & \textbf{0.137} & \textbf{0.620} & \textbf{2.04} \\
\bottomrule
\end{tabular}
\end{table}

\subsection{Comparison with LIFe-GoM}
We additionally compare the proposed method with LIFe-GoM~\cite{lifegom}, which also addresses generalizable human rendering from sparse-view inputs, on the THuman2.0 dataset~\cite{thuman}.
To provide a fair comparison, we follow the evaluation protocol of LIFe-GoM.
Both the source image and target rendering resolutions are set to $1024\times1024$, and a $500\times1000$ region centered on the human subject is cropped from both the rendered and ground-truth images for evaluation.
We report PSNR, LPIPS$^{*}$ (i.e., LPIPS$\times1000$), and FID, where FID evaluates the distributional similarity between the rendered and ground truth image sets.
As shown in Table~\ref{tab:lifegom}, the proposed method substantially outperforms LIFe-GoM under its evaluation protocol.
Specifically, the proposed method improves PSNR by $4.12$~dB and reduces LPIPS$^{*}$ by $59.31$, while achieving $7.7$ FPS compared with $1.0$ FPS of LIFe-GoM.
These results demonstrate that the proposed method provides both superior rendering quality and faster inference speed.

\begin{table}[h]
\centering
\vspace{3px}
\caption{\textbf{Quantitative comparison with LIFe-GoM~\cite{lifegom} on the THuman2.0 dataset.} The best result is shown in bold.}
\label{tab:lifegom}
\footnotesize
\setlength{\tabcolsep}{8pt}
\begin{tabular}{lcccc}
\toprule
Methods & PSNR$\uparrow$ & LPIPS$^{*}\downarrow$ & FID$\downarrow$ & FPS$\uparrow$ \\
\midrule
LIFe-GoM~\cite{lifegom} & 25.10 & 108.75 & 45.17 & 1.0 \\
\textbf{Ours} & \textbf{29.22} & \textbf{49.44} & \textbf{7.91} & \textbf{7.7} \\
\bottomrule
\end{tabular}
\vspace{3px}
\end{table}

\subsection{Comparison with single-image human reconstruction methods}
Recent large reconstruction models, i.e., IDOL~\cite{idol} and LHM~\cite{lhm}, reconstruct animatable human avatars from a single image, whereas the proposed method focuses on generalizable human Gaussian splatting from multiple calibrated views.
We compare the proposed method with IDOL on the THuman2.1 dataset~\cite{thuman} and with LHM on the 2K2K dataset~\cite{2k2k}, where each comparison follows the experimental setting of the corresponding method.
As shown in Table~\ref{tab:single_image}, the proposed method achieves better performance across all metrics, since it takes four input views while IDOL and LHM take a single image.
Note that the results of the proposed method in Table~\ref{tab:single_image} differ from those in Tables~\ref{tab:in_domain_others} and~\ref{tab:thuman21} due to the difference in evaluation protocols.

\begin{table}[h]
\centering
\caption{\textbf{Quantitative comparison with single-image human reconstruction methods.} The best result is shown in bold.}
\label{tab:single_image}
\footnotesize
\setlength{\tabcolsep}{8pt}
\begin{tabular}{llcccc}
\toprule
Datasets & Methods & \# of input views & PSNR$\uparrow$ & SSIM$\uparrow$ & LPIPS$\downarrow$ \\
\midrule
\multirow{2}{*}{THuman2.1} & IDOL~\cite{idol} & 1 & 22.12 & 0.9257 & 0.112 \\
 & \textbf{Ours} & 4 & \textbf{33.51} & \textbf{0.9802} & \textbf{0.017} \\
\midrule
\multirow{2}{*}{2K2K} & LHM~\cite{lhm} & 1 & 20.73 & 0.9084 & 0.122 \\
 & \textbf{Ours} & 4 & \textbf{31.72} & \textbf{0.9722} & \textbf{0.022} \\
\bottomrule
\end{tabular}
\end{table}

\section{Limitations and future directions}
\label{sec:limitations}
\noindent\textbf{Static reconstruction.}
The proposed method focuses on novel view synthesis from a static set of multi-view images, which does not explicitly handle dynamic scenarios.
Consequently, applying the proposed method to video sequences would require per-frame reconstruction, which is computationally expensive and may produce temporally inconsistent results across frames.
A promising direction to address this limitation is to incorporate a parametric human body model, e.g., SMPL~\cite{smpl}, into the proposed framework, where the predicted Gaussians can be deformed according to a temporal sequence of target poses via linear blend skinning.
Extending the proposed method along this direction remains as the future work.

\noindent\textbf{Dependency on the clothed human mesh.}
Since the complexity prior is computed from the clothed human mesh, our method requires an accurate clothed human mesh during the training phase.
In practice, obtaining accurate clothed human meshes remains challenging, as it typically relies on labor-intensive 3D scanning.
A promising direction to address this limitation is to investigate complexity priors directly computed from multi-view input images.
This would reduce the dependency on the clothed human mesh during training and broaden the range of training datasets.

\section{Ethics statement}
\label{sec:ethics}
We use publicly available human datasets, including THuman~\cite{thuman}, 2K2K~\cite{2k2k}, X-Humans~\cite{xhuman}, DNA-Rendering~\cite{dna}, and ZJU-MoCap~\cite{zju}, all of which provide properly released data for research purposes.
Our research focuses on generalizable human Gaussian splatting from sparse multi-view images, a technology that can benefit applications in graphics, telepresence, virtual reality, and gaming.
However, as this technique enables the reconstruction of a realistic human appearance from visual observations, it may raise concerns related to privacy, identity misuse, or unauthorized replication of personal likeness.
We emphasize that the proposed method is intended solely for research use and should not be applied to scenarios involving unauthorized digital reproduction of individuals.
Responsible deployment and adherence to ethical guidelines are strongly encouraged.

\clearpage
\section*{NeurIPS Paper Checklist}

\begin{enumerate}

\item {\bf Claims}
    \item[] Question: Do the main claims made in the abstract and introduction accurately reflect the paper's contributions and scope?
    \item[] Answer: \answerYes{} 
    \item[] Justification: The claims in the abstract and the introduction match the main contributions.

\item {\bf Limitations}
    \item[] Question: Does the paper discuss the limitations of the work performed by the authors?
    \item[] Answer: \answerYes{} 
    \item[] Justification: We discuss the limitations and possible solutions in the supplementary material due to page limits.

\item {\bf Theory assumptions and proofs}
    \item[] Question: For each theoretical result, does the paper provide the full set of assumptions and a complete (and correct) proof?
    \item[] Answer: \answerNA{} 
    \item[] Justification: The paper does not include theoretical results.

    \item {\bf Experimental result reproducibility}
    \item[] Question: Does the paper fully disclose all the information needed to reproduce the main experimental results of the paper to the extent that it affects the main claims and/or conclusions of the paper (regardless of whether the code and data are provided or not)?
    \item[] Answer: \answerYes{} 
    \item[] Justification: We provide essential implementation details in the manuscript and the supplementary material.

\item {\bf Open access to data and code}
    \item[] Question: Does the paper provide open access to the data and code, with sufficient instructions to faithfully reproduce the main experimental results, as described in supplemental material?
    \item[] Answer: \answerYes{} 
    \item[] Justification:  The code and pretrained model are publicly available at the link provided in the abstract.

\item {\bf Experimental setting/details}
    \item[] Question: Does the paper specify all the training and test details (e.g., data splits, hyperparameters, how they were chosen, type of optimizer) necessary to understand the results?
    \item[] Answer: \answerYes{} 
    \item[] Justification: We describe the detailed experimental settings in the manuscript and the
supplementary material.

\item {\bf Experiment statistical significance}
    \item[] Question: Does the paper report error bars suitably and correctly defined or other appropriate information about the statistical significance of the experiments?
    \item[] Answer: \answerNo{} 
    \item[] Justification: The results are obtained from standard randomly seeded runs without cherrypicking, offering a fair and representative evaluation despite not including error bars.

\item {\bf Experiments compute resources}
    \item[] Question: For each experiment, does the paper provide sufficient information on the computer resources (type of compute workers, memory, time of execution) needed to reproduce the experiments?
    \item[] Answer: \answerYes{} 
    \item[] Justification: The paper specifies the type of GPU used, memory, and time of execution needed to reproduce the experiments in the manuscript and the supplementary material.
    
\item {\bf Code of ethics}
    \item[] Question: Does the research conducted in the paper conform, in every respect, with the NeurIPS Code of Ethics \url{https://neurips.cc/public/EthicsGuidelines}?
    \item[] Answer: \answerYes{} 
    \item[] Justification: The research adheres to the NeurIPS Code of Ethics, with no use of sensitive
data or methods posing foreseeable ethical risks.

\item {\bf Broader impacts}
    \item[] Question: Does the paper discuss both potential positive societal impacts and negative societal impacts of the work performed?
    \item[] Answer: \answerYes{} 
    \item[] Justification: The paper discusses  broader impacts in the manuscript.
    
\item {\bf Safeguards}
    \item[] Question: Does the paper describe safeguards that have been put in place for responsible release of data or models that have a high risk for misuse (e.g., pre-trained language models, image generators, or scraped datasets)?
    \item[] Answer: \answerNA{} 
    \item[] Justification: The work does not involve releasing models or data with high risk of misuse,
so specific safeguards are not applicable.

\item {\bf Licenses for existing assets}
    \item[] Question: Are the creators or original owners of assets (e.g., code, data, models), used in the paper, properly credited and are the license and terms of use explicitly mentioned and properly respected?
    \item[] Answer: \answerYes{} 
    \item[] Justification: All external datasets and code-bases used in the paper are properly cited.

\item {\bf New assets}
    \item[] Question: Are new assets introduced in the paper well documented and is the documentation provided alongside the assets?
    \item[] Answer: \answerNA{} 
    \item[] Justification: The paper does not introduce new assets such as datasets that require separate
documentation.

\item {\bf Crowdsourcing and research with human subjects}
    \item[] Question: For crowdsourcing experiments and research with human subjects, does the paper include the full text of instructions given to participants and screenshots, if applicable, as well as details about compensation (if any)? 
    \item[] Answer: \answerNA{} 
    \item[] Justification: We mainly utilize publicly available datasets for our experiments.

\item {\bf Institutional review board (IRB) approvals or equivalent for research with human subjects}
    \item[] Question: Does the paper describe potential risks incurred by study participants, whether such risks were disclosed to the subjects, and whether Institutional Review Board (IRB) approvals (or an equivalent approval/review based on the requirements of your country or institution) were obtained?
    \item[] Answer: \answerNA{} 
    \item[] Justification: The paper does not involve crowdsourcing nor research with human subjects.

\item {\bf Declaration of LLM usage}
    \item[] Question: Does the paper describe the usage of LLMs if it is an important, original, or non-standard component of the core methods in this research? Note that if the LLM is used only for writing, editing, or formatting purposes and does \emph{not} impact the core methodology, scientific rigor, or originality of the research, declaration is not required.
    \item[] Answer: \answerNA{} 
    \item[] Justification: The core method development in this research does not involve LLMs as any
important, original, or non-standard components.

\end{enumerate}
\end{document}